\documentclass{article}

     \usepackage[final]{neurips_2020}

\usepackage[utf8]{inputenc} 
\usepackage[T1]{fontenc}    
\usepackage{hyperref}       
\usepackage{url}            
\usepackage{booktabs}       
\usepackage{amsfonts}       
\usepackage{nicefrac}       
\usepackage{microtype}      

\usepackage[square,sort,comma,numbers]{natbib}

\usepackage{amsmath}
\usepackage{amssymb}
\usepackage{times}
\usepackage{soul}
\usepackage{url}
\usepackage{amsmath}
\usepackage{booktabs}
\usepackage{array}
\usepackage{multirow}
\usepackage{multicol}
\usepackage{tabu}
\usepackage{epsfig}
\usepackage{graphicx}
\usepackage[title]{appendix}
\usepackage{amsmath}
\usepackage{amssymb}
\usepackage{amsfonts}
\usepackage{color}
\usepackage{colortbl}
\usepackage{epsfig}
\usepackage{multirow}
\usepackage{algorithm}
\usepackage{algpseudocode}
\usepackage{scrextend}
\usepackage{tabularx}
\usepackage{bbm}
\usepackage{graphics}
\usepackage{color}
\usepackage{wrapfig}
\usepackage{footnote}
\usepackage{bm}
\usepackage[autolanguage,boldmath]{numprint}
\usepackage[font=small,skip=0pt,belowskip=0pt]{caption}
\usepackage{subcaption}

\usepackage[dvipsnames]{xcolor}
\usepackage{tikz}
\usepackage{pgfplots}

\usepackage{comment}

\npdecimalsign{.}
\nprounddigits{3}

\newcommand{\topscore}{{\fontseries{b}\selectfont}}
\newcommand{\topRed}{{\color{red}}}
\newcommand{\topBRed}{{\fontseries{b}\selectfont \color{red}}}

\usepackage{titlesec}

\titlespacing\section{0pt}{12pt plus 4pt minus 2pt}{0pt plus 2pt minus 2pt}
\titlespacing\subsection{0pt}{12pt plus 4pt minus 2pt}{0pt plus 2pt minus 2pt}
\titlespacing\subsubsection{0pt}{12pt plus 4pt minus 2pt}{0pt plus 2pt minus 2pt}

\title{Private-Shared Disentangled Multimodal VAE for Learning of Hybrid Latent Representations}

\author{%
  Mihee Lee \\
  Department of Computer Science\\
  Rutgers University, Piscataway, NJ, USA\\
  \texttt{ml1323@rutgers.edu} \\
  \AND
  Vladimir Pavlovic \\
  Department of Computer Science\\
  Rutgers University, Piscataway, NJ, USA\\
  \texttt{vladimir@cs.rutgers.edu} \\
}

\begin{document}

\setlength{\abovedisplayskip}{3pt}
\setlength{\belowdisplayskip}{3pt}

\maketitle

\begin{abstract}
Multi-modal generative models represent an important family of deep models, whose goal is to facilitate representation learning on data with multiple views or modalities.  However, current deep multi-modal models focus on the inference of shared representations, while neglecting the important private aspects of data within individual modalities. In this paper, we introduce a disentangled multi-modal variational autoencoder (DMVAE) that utilizes disentangled VAE strategy to separate the private and shared latent spaces of multiple modalities. We specifically consider the instance where the latent factor may be of both continuous and discrete nature, leading to the family of general hybrid DMVAE models.  We demonstrate the utility of DMVAE on a semi-supervised learning task, where one of the modalities contains partial data labels, both relevant and irrelevant to the other modality. Our experiments on several benchmarks
indicate the importance of the private-shared disentanglement as well as the hybrid latent representation.
\end{abstract}

\section{Introduction}
Representation learning is a key step in the process of data understanding, where the goal is to distill interpretable factors associated with the data. However, typically representation learning approaches focus on data observed in a single modality, such as text, images, or video. Nevertheless, most real world data comes from processes that manifest itself in multiple views or modalities.  In computer vision, image-based data is typically accompanied with strong or weak labels summarized in data attributes.  
For example, 
an image of a smiling woman with eye glasses is augmented with labels describing both visual and non-visual attributes of that individual.  Visual attributes, such as the woman displaying open mouth, smiling, and having eyeglasses, have their counterparts in the image domain.  However, other attributes such as the woman's marital status, job, etc., cannot be deduced from the image view.  Similarly, many image attributes are not captured in the accompanied labels.  Therefore, accurate modeling of the underlying data representation has to consider both the \textbf{private} aspects of individual modalities as well as what those modalities \textbf{share}. 


 

In this paper, we propose a generative variational model that can learn both the private and the shared latent space of each modality, with each latent variable attributed to a disentangled representational factor.  The model extends the well-known family of  Variational AutoEncoders (VAEs)~\cite{KingmaW13} by introducing separate shared and private spaces, whose representations are induced using pairs of individual modality encoders and decoders.  To create the shared representation, we impose consistency of representations using a product-of-experts (PoE)~\cite{819532} inference network. The key requirement for separation of representations among the private and shared spaces is imposed using the disentanglement criteria, traditionally used in single-modal representations~\cite{pmlr-v80-kim18b,chen2018isolating,esmaeili2018structured}.  
Finally, we tackle another critical issue that arises in many data representation learning approaches, where the latent factors may exhibit both continuous and discrete (categorical) nature. We address this issue by modeling the discrete factors using the Gumbel SoftMax approximation~\cite{MaddisonMT17,dupont2018learning} and the continuous factors using a homoscedastic normal model. This representation is effectively combined in an end-to-end learning framework with the private-shared disentangled VAE, resulting in the novel disentangled multi-modal variational autoencoder (DMVAE).

We apply DMVAE to two multi-modal representation learning problems.  In the first setting, we consider the problem of learning the shared/private generative representations of digit images, where the accompanying modality is the digit class.  In the second setting, we aim to model the representation of facial images and their multiple attributes.  In both instances, the class labels and the attributes are not consistently paired with image data points, leading to a semi-supervised learning setting.
We contrast our approach with MVAE~\cite{NIPS2018_7801} model, which relies on a single shared continuous representation without private spaces.

\section{Related Work}
Our work addresses two aspects of representation modeling, the disentangled representations and the multi-modal data.  We consider relevant prior work in these two contexts.

\noindent\textbf{Disentangled representations.} A number of approaches have been suggested to reveal disentangled factors of a given datasets. InfoGAN \cite{NIPS2016_6399} achieves disentangled representations by maximizing the mutual information between a subset of the latents and the input data based on the Generative Adversarial Network(GAN) \cite{NIPS2014_5423}. FactorVAE \cite{pmlr-v80-kim18b} leverages total correlation (TC) to control the dependency between each of the latents in the Kullback–Leibler (KL) divergence term of the VAE objective. It also penalizes TC to induce independence among latent factors. FactorVAE requires an additional training phase for the discriminator as well as the training of the VAE model to achieve the disentanglement. $\beta$-TCVAE \cite{chen2018isolating} suggests an insightful decomposition of the KL, breaking it down into Mutual Information (MI) between the input data and its latents, TC across all latents, and the distance between the empirical and the prior distribution of each latent variable. They stochastically estimate each of these three terms using minibatch-weighted sampling based on Monte Carlo approximation, eliminating the need for additional training of a discriminator in FactorVAE. HFVAE \cite{esmaeili2018structured} investigates a two-level hierarchical factorization of the VAE by grouping relevant factors. By doing so, HFVAE models the relation between groups as well as the individual factor within a group. Moreover, HFVAE utilizes discrete latents as in \cite{dupont2018learning} and jointly learns the continuous-discrete representation.  While all prior works attempt to infer disentangled representations in an unsupervised setting, to the best of our knowledge, our DMVAE represents a first attempt to tie multi-modality with disentanglement in order to model the critical but disjoint private and shared space in a multi-modal setting.

\begin{figure}[tbhp]
\centering
 \begin{subfigure}[b]{0.24\textwidth}
 {\includegraphics[width=\textwidth]{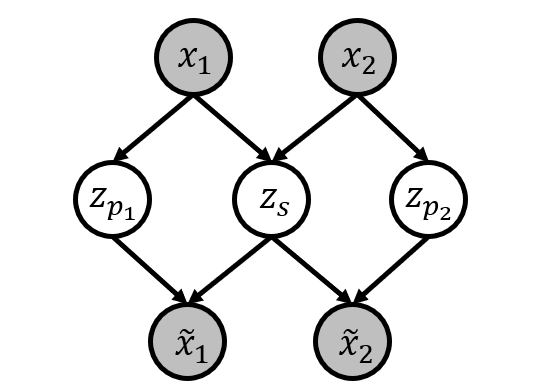} \caption{Full-modality}\label{fig:full} }
 \end{subfigure}
 \hfill
 \begin{subfigure}[b]{0.24\textwidth} 
 {\includegraphics[width=\textwidth]{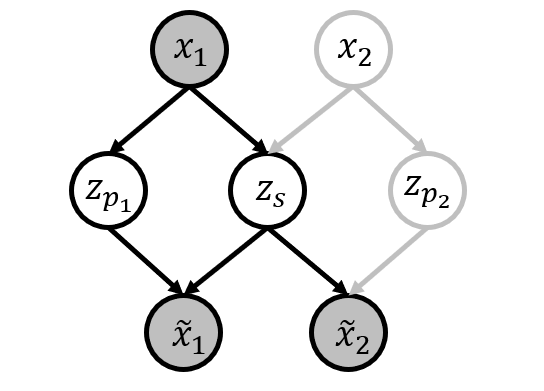} \caption{Missing-modality} \label{fig:missing}}
 \end{subfigure}    
 \hfill
\begin{subfigure}[b]{0.45\textwidth} 
{\includegraphics[width=\textwidth]{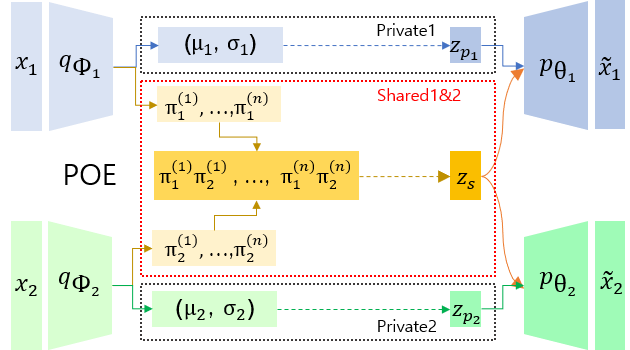} \caption{Model architecture}\label{fig:arch_disc}}
\end{subfigure}

  \caption{(a-b) Unrolled graphical model representation of DMVAE.  The gray circles illustrate observed variables. ${z_{p_1}}, {z_{p_2}}$ denote the private latents of modalities ${x_1}, {x_2}$. ${z_s}$ denotes the shared latents between two modalities. ${\tilde{x}_1}, {\tilde{x}_2}$ denote reconstructed views, which should match the observed data in this unrolled generative model. (b) illustrates the missing modality instance network, which is critical for test-time inference of $x_2$ from $x_1$. We elaborate the inference in missing modality in \autoref{sec4.infer1}. (c) Model architecture of DMVAE. Each modality infers the shared latent variable and then those shared spaces from multi-modalities are aligned by product-of-expert. The dashed line indicates sampling from a distribution.}
\label{fig:graph}
\vspace{-.9em}
\end{figure}

\noindent\textbf{Multi-modal Learning.} Several prior works have considered the problem of modeling multi-modal data using generative VAE-inspired models. 
JMVAE \cite{suzuki2016joint} exploits the joint inference network $q(z|x_1,x_2)$ to learn the interaction of two modalities, $x_1$ and $x_2$. To address the missing modality problem, where some of the data samples are not paired (i.e., do not have both views present), they train inference networks $q(z|x_1)$, $q(z|x_2)$ in addition to the bimodal inference model $q(z|x_1,x_2)$, and then minimize the distance between uni- and multi-modal based latent distribution.
JVAE \cite{VedantamFH018} adopts a product-of-expert (PoE) \cite{819532} for the joint posterior $q(z|x_1,x_2)$ of multi modalities in the inference network. The approach leverages the unimodal inference networks, whose predictions are constrained and made consistent through the PoE. JVAE trains the model with two-stage process to handle both paired and missing modality data.  Due to this fact, the number of required inference networks increases exponentially for more than two modalities. To alleviate the inefficiency of JVAE, MVAE \cite{NIPS2018_7801} considers only partial combination of observed modalities. This helps reduce the number of parameters and  increase the computational efficiency of learning.
\cite {Shi2019VariationalMA} applies Mixture-of-Expert (MoE) to jointly learn the shared factors across multi-modalities. Though they introduces the concept of the private and shared information of multi-modalities, it is implicitly conceived. Moreover, the use IWAE for the approximation makes the contirbution of MoE vague.

However, all of the mentioned prior works use a single latent space to represent the multi-modal data.  Within this common latent space, modality-specific factors could be entangled with the shared factors across all moralities, reducing the ability of these generative models to represent the data and infer the "true" latent factors. In this paper, we solve this problem by explicitly separating the shared from the disjoint private spaces, using individual inference networks to achieve this goal, coupled with the disentanglement criteria. This is illustrated in \autoref{fig:graph} and elaborated on in subsequent sections.

\section{\vspace{-.3em}Background}

Our DMVAE framework builds upon the VAE model of \cite{KingmaW13}.  We first highlight the relevant aspects of VAE-based models, which we then leverage to construct the DMVAE in \autoref{sec:dmvae_method}.

\noindent\textbf{Variational Autoencoder.} A variational autoencoder (VAE) \cite{KingmaW13} implements variational inference for the latent variable via autoencoder structure. The objective of the VAE is to maximize the marginal distribution
$ p(x) =  \int p_{\theta}(x|z)p(z)dx$ 
which is, however, intractable. 
Thus, VAE introduces the evidence lower bound (ELBO) which uses an approximated recognition model $q_{\phi}(z|x)$ instead of the intractable true posterior. It maximizes ${\mathbb{E}}_{z {\sim} q_{\phi}(z|x)}\left[{\log{p_{\theta}(x|z)}} \right]$ while minimizing KL($q_{\phi}(z|x), p(z)$). 
\begin{equation}\label{eq:elbo}
\begin{split}
\log{p(x)} & \geq {\mathbb{E}}_{p(x)}\left[ELBO(x;\theta, \phi)\right] \\
&={\mathbb{E}}_{p(x)}\left[{\mathbb{E}}_{q_{\phi}(z|x)}\left[{\log{p_{\theta}(x|z)}} \right] - {\beta}KL(q_{\phi}(z|x)||p(z)) \right] 
\end{split}
\end{equation}
where ${\beta} = 1$.
$q_{\phi}(z|x)$ and $p_{\theta}(x|z)$ are represented as encoder and decoder in the network with the learning parameter ${\phi}$ and ${\theta}$, respectively. The first term of ELBO (Eq. \eqref{eq:elbo}) is the reconstruction error and the second term plays a role of regularizer not to be far from the prior distribution $p(z)$. We next discuss in more detail the effect of the second term on disentanglement.

\noindent\textbf{ELBO KL-Decomposition.} $\beta$-VAE \cite{Higgins2017betaVAELB} emphasizes the KL term in ELBO with an amplification factor ${\beta} > 1$, demonstrating its ability to improve the disentangling of latent factors. However, this inevitably leads to degradation in data reconstruction. To relieve this problem, FactorVAE~\cite{pmlr-v80-kim18b} and $\beta$-TCVAE~\cite{chen2018isolating} explicitly decompose the KL divergence between the variational posterior $q_{\phi}(z|x)$ and the prior $ p(z)$. They control each component of the KL term considering the trade-off with reconstruction quality;
$\beta$-TCVAE~\cite{chen2018isolating} breaks down the KL term of ELBO as:
\begin{multline}\label{eq:betatc}
{\mathbb{E}}_{p(x)}\left[KL(q_{\phi}(z|x)||p(z)) \right] 
= \overbrace{KL(q_{\phi}(z,x)||q_{\phi}(z)p(x))}^{MI(x;z)} \\ 
+ \overbrace{KL(q_{\phi}(z)||\prod_{k}{q_{\phi}(z_k)})}^{TC(z;\prod_{k}{q_{\phi}(z_k)})} + \overbrace{\sum_{k}{KL(q_{\phi}(z_k)||p(z_k))}}^{FP(q_{\phi},p)},
\end{multline}
where the first term, $MI(x;z)$, encodes dependency between the latent representation and the input.  The second term, $TC(z;\prod_{k}{q_{\phi}(z_k)})$, encourages independence between latent representations in individual factors, and is directly related to disentanglement. Finally, the last term, $FP(q_{\phi},p)$, regularizes the factorized latent representation with respect to the prior. 

$\beta$-TCVAE~\cite{chen2018isolating} derives an analytic formula to estimate these terms stochastically with minibatch-weighted sampling based on Monte Carlo approximation. In this paper, we focus on the $TC$ term to encourage disentanglement of private and shared factors.


\section{DMVAE Framework}\label{sec:dmvae_method}
In this section, we introduce the new DMVAE model. 
\autoref{sec4.ps} describes the architecture of private and shared latent spaces within the disentangled representation. 
In \autoref{sec4.infer1} and \autoref{sec4.infer2},
we define the DMVAE inference models, accompanied with the learning objective in \autoref{sec4.obj}.
 \autoref{sec4.hybrid} explains how the PoE of categorical distributions is derived to join the shared spaces across multiple modalities.

\subsection{Private / Shared-Disentangled Multi-Modal VAE}  \label{sec4.ps}

The assumption in this paper is that under the multi-modal description of a concept, the latent space of the concept is divided into a private space of each modality and one shared space across all modalities\footnote{Other more intricate representations of private and shared spaces may arise in the presence of more than two modalities. However, we do not consider this setting in our current work.}. 
Our goal is to obtain well-separated private and shared spaces. This separation is critical;  the shared latent space can only transfer the information common across modalities, but it will fail to model the individual aspects of the modalities. In a generative model, such as the VAE, modeling the private factors is critical as those factors enable both the high fidelity of the data reconstruction as well as the improved separation (disentanglement) of the latent factors across modalities.

Our model is illustrated in \autoref{fig:arch_disc} for the case of two modalities. Given  paired \textit{i.i.d.} data $\{(x_1, x_2)\}$, we infer the latents $z_1 \sim q_{\phi_1}(z|x_1), z_2 \sim q_{\phi_2}(z|x_2)$, where ${\phi_1}, {\phi_2}$ are the parameters of each individual modal inference network. We assume the latents can be factorized into $z_1 = [z_{p_1}, z_{s_1}]$ and $z_2 = [z_{p_2}, z_{s_2}]$, where ${z_{p_1}}, {z_{p_2}}$ represent the private latents of modalities ${x_1}, {x_2}$, respectively, and ${z_{s_1}}, {z_{s_2}}$ represent the shared latents, which are to model the commonality between the two modalities.

For the desired shared representation in ${z_{s_1}}, {z_{s_2}}$ we seek to effectively make ${z_{s_1}} = {z_{s_2}}$. We describe how to accomplish this using a PoE-based consistency model in \autoref{sec4.infer1}, which approximates the shared inference network $p(z_s|x_1, x_2)$. Furthermore, to encourage separation among private and shared factors, we leverage the disentanglement approach by focusing on the total correlation $TC(z;\prod_{k}{q_{\phi}(z_k)})$ of each latent factor. We penalize the second term in \eqref{eq:betatc} to impose independence (statistical orthogonality) of each $z_i$ in $z$ from other private and shared factors. 
Thus, our model can both learn the interpretable factors and achieve segregation of private and shared latent spaces.


\subsection{Latent Space Inference}\label{sec4.infer1}
First, we define the latent space inference in our model. Given $N$ modalities $\bm{x} = (x_1, \ldots, x_N)$, each modality has the posterior distribution $p(z_i|x_i)$, approximated by inference networks $q(z_i|x_i) = q(z_{p_i}, z_{s_i}|x_i)$. Since the shared latent space should reflect the information shared across all modalities, we require that the representation be consistent, i.e., $z_{s_i}=z_{s}~\text{w.p.}1, \forall i$. Consequently, we separate the \textit{private inference} $q(z_{p_i}|x_i)$ from the \textit{shared inference} network $q(z_{s}|\bm{x})$, defined using the product-of-experts (PoE) model~\cite{819532}, adopted in \cite{VedantamFH018, NIPS2018_7801}:
\begin{equation}\label{eq:poe}
 q(z_s|\bm{x}) \propto p(z_s) {\prod_{i=1}^{N} {q(z_s | x_{i})}} 
\end{equation} 
In the case where all inference networks and priors assume conditional Gaussian forms,  $p(z) = \mathcal{N}(z|0,I)$ and $q(z|x_i) = \mathcal{N}(z|\mu_i, C_i)$ of i-th Gaussian expert with the covariance $C_i$, the PoE shared inference network will have the closed form of $q(z|\bm{x})$ as $\mathcal{N}(z|\mu, C)$ where $C^{-1} = \sum_i{C_i^{-1}}$ and $\mu = C \sum_i{C_i^{-1}}\mu_i$.

\noindent\textbf{Missing-mode Inference.} An important benefit of the PoE-induced shared inference is that the individual modality shared networks can also be used for inference in instances when one (or more) of the modalities is missing.  Specifically, as illustrated in the bi-modal case of \autoref{fig:missing}, under the $x_2$ missing, the shared latent space would be simply inferred using the remaining modality shared inference network $q(z_s|x_1)$; and vice-versa for missing $x_1$.

\subsection{Reconstruction Inference}\label{sec4.infer2}
In addition to inferring the latent factors, a key enabler in VAE is the reconstruction inference, or encoding-decoding.  Specifically, we seek to infer 
$p(\tilde{x} | x) = \int p(\tilde{x},z | x) dz = \int p(\tilde{x} | z) p(z | x) dz = {\mathbb{E}}_{p(z|x)}[p({\tilde{x}}|z)] \approx {\mathbb{E}}_{q(z|x)}[p({\tilde{x}}|z)]$.

The reconstruction inference in multi-modal settings, much like the latent space inference, has to consider the cases of complete and missing modality data.  We assume bi-modality without loss of generality. The first case is the self-reconstruction within a single modality, $\mathbb{E}_{q(z_{p_i}|x_i)q(z_s|x_i)} [p({\tilde{x}_i}|z_{p_i}, z_s)]$ for $i=1,2$. The second form is the joint multi-modal reconstruction, ${\mathbb{E}}_{q(z_{p_i}|x_i)q(z_s|x_1, x_2)}[p({\tilde{x}_i}|z_{p_i}, z_s)]$ for $i=1,2$. It is also possible to consider the cross-modal reconstruction, e.g., 
$p({\tilde{x}}_2|x_1) = {\mathbb{E}}_{p(z_{p_2}) q(z_s|x_1)}[p({\tilde{x}}_2|z_{p_2}, z_s)]$, illustrated in \autoref{fig:missing}.  This instance, where $x_2$ is missing, is facilitated using the prior on the private space of $x_2$, $p(z_{p_2})$.

The different reconstruction inference modes are essential for model learning but also valuable for understanding the model performance.  For instance, one may seek to see how successful the multi-modal DMVAE is in learning the shared and private representations in the context of synthesizing one modality from another.  We highlight these cross-synthesis experiments in \autoref{sec5.single} and \autoref{sec5.multi}.

\subsection{Learning Objective} \label{sec4.obj}

Learning the DMVAE model follows the general $\beta$-TCVAE paradigm, generalized to the multi-modal setting.  However, a multimodal setting benefits from both the fully paired training data, $(x_1,x_2)$ in the bimodal setting, as well as the unimodal instances, $x_1$ or $x_2$, with the other modality missing.

In general, for each data point $\bm{x}$ and $N$ modalities, the learning objective assumes the form:
\begin{multline}\label{eq:objective}
\sum_i \mathbb{E}_{p(x_i)}\biggl[ {\lambda_i}{\mathbb{E}}_{q_{\phi}(z_{p_i},|x_i), q_{\phi}(z_{s},|\bm{x})}\left[{\log{p_{\theta}(x_i|z_{p_i},z_s)}} \right]
- KL(q_{\phi}(z_{p_i}|x_i)||p(z_{p_i})) - KL(q_{\phi}(z_s|\bm{x})||p(z_s)) \\
+ \sum_j \Big({\lambda_i}{\mathbb{E}}_{q_{\phi}(z_{p_i},|x_i), q_{\phi}(z_{s},|x_j)}\left[{\log{p_{\theta}(x_i|z_{p_i},z_s)}} \right] 
- KL(q_{\phi}(z_{p_i}|x_i)||p(z_{p_i}))
- KL(q_{\phi}(z_s|x_j)||p(z_s))\Big)\biggr],
\end{multline}

where $\lambda_i$ is balancing reconstruction across different modalities. The first term models the accuracy of reconstruction with the jointly learned shared latent factor, compensated by the KL-divergence from the prior.  The second set of terms assesses the accuracy of the cross-modal reconstruction, $x_i \leftarrow x_j$ for i $\neq j$ and the accuracy of self-reconstruction for $i = j$ , again compensated by the divergence. 

We further decompose each KL term by applying $\beta$-TCVAE \cite{chen2018isolating} decomposition as in Eq.~\eqref{eq:betatc}, to control the disentanglement between latents, essential for private-shared separation, as well as use the regularization from assumed priors. Specifically, each KL term in Eq.~\eqref{eq:objective} is explicitly assumed to be: 
\begin{multline}
{\mathbb{E}}_{p(x_i)}\left[KL(q_{\phi}(z|x_i)||p(z)) \right] \\
= KL(q_{\phi}(z,x_i)||q_{\phi}(z)p(x_i)) + \boldsymbol{\beta}_i KL(q_{\phi}(z)||\prod_{k}{q_{\phi}(z_k)}) + \sum_{k}{KL(q_{\phi}(z_k)||p(z_k))},
\end{multline}
where $z$ can be either $z_s$ or $z_{p_i}$.

\subsection{Hybrid Posterior for Continuous / Discrete Shared Latent Space} \label{sec4.hybrid}

In addition to continuous shared space with a Gaussian distribution, this paper considers discrete shared space with a \textit{categorical distribution}. Categorical latent factors are essential for models that deal with categorical attributes as one modality, which is the setting we consider in our work.  However, dealing with categorical latents is challenging because of typically non-differentiable objectives that arise from such latents.

To make the discrete variable differentiable, we adopt a concrete distribution using the Gumbel Max trick as a relaxation of the underlying discrete distribution. Here, we assume two modalities $x_1, x_2$ to simplify notation, wlog. For $G_k \sim$ Gumbel(0,1) \textit{i.i.d.} and temperature $T \in (0, \infty)$, a concrete random variable $z_{s_i} \in \Delta^{n-1} = \{z_{s_i} \in \mathbb{R}^n | z_{s_i}^{(j)} \in [0,1], \sum_{j=1}^n{z_{s_i}^{(j)}}=1 \}$ is sampled as follows.
\begin{equation}\label{eq:concrete}
 z_{s_i}^{(k)} = \nicefrac {\exp\frac{1}{T}(\log{\pi_{i}^{(k)}} + G_k)} {\sum_{j=1}^n{\exp\frac{1}{T}(\log{\pi_{i}^{(j)}} + G_j)}}
\end{equation}
where $i \in \{1, 2\}$ is the index for modalities, $n$ is the dimension of the shared space,  $\pi_{i}^{(1)}, \pi_{i}^{(2)}, \dots , \pi_{i}^{(n)}$ are class probabilities for $z_{s_i} \in \mathbb{R}^n$, the output of our inference network as illustrated in \autoref{fig:arch_disc}. By replacing argmax of categorical variable with softmax and using the reparametrization trick with $G_k = - \log(-\log(u_k))$ where $u_k \sim$ Unif(0,1), a random variable $z_{s_i}$ becomes differentiable. As the temperature $T \rightarrow 0$ , this relaxation gets closer to the original categorical distribution. A concrete random variable $z_{s_i} \sim $ Concrete($\pi_{i}, T$) has its density function 
\begin{equation}\label{eq:concrete_density}
 p_{\pi_{i},T}(z_{s_i}) = (n-1)! T^{n-1} \prod_{k=1}^n\nicefrac{\pi_{i}^{(k)} / (z_{s_i}^{(k)})^{T+1}} {\sum_{j=1}^n{\pi_{i}^{(j)} / (z_{s_i}^{(j)})^{T}}}
\end{equation} 

We set the prior as a concrete relaxation of Uniform distribution with the same class probability $1/n$ by assuming that the class probability is all equal across all classes. Thus, the prior is $z_s \sim $ Concrete($1/n, T$) and each modality has the posterior distribution $z_{s_1} \sim $ Concrete($\pi_{1}, T$) and $z_{s_2} \sim $ Concrete($\pi_{2}, T$). From Eq. \eqref{eq:poe}, the joint posterior of concrete variables by PoE is
\begin{equation}
\begin{split}
\label{eq:concrete_poe}
 q(z|x_1, x_2) & \propto p(z) q(z|x_1) q(z|x_2)  = p(z) p_{\pi_{1},T}(z) p_{\pi_{2},T}(z) \\
&\approx ((n-1)! T^{n-1})^3 \prod_{k=1}^n\frac{\pi_{1}^{(k)}\pi_{2}^{(k)} ((z_k^3)^{-T-1})} {\sum_{j=1}^n{\pi_{1}^{(j)}\pi_{2}^{(j)} (z_j^3)^{-T}}} \propto p_{\pi_{1}\pi_{2},T}(z^3).
\end{split}
\end{equation}

The approximation holds by the property of concrete distribution on the simplex and the assumption that the multi-modal data is given as a pair. First, the argmax computation of the categorical distribution has the property that it returns states on the vertices of the simplex $\Delta^{n-1}$. \footnote{We illustrate how PoE and true concrete distributions change as a function of $T$ in the Supplement.} Because the concrete distribution approximates the categorical distribution very closely as $T \rightarrow 0$, it can also satisfy the desired property.  Furthermore, the paired input from all modalities places the same class information in the shared space, which completes $z_iz_j \approx 0 $ for $i \neq j$ as $T \rightarrow 0$.

\section{Experiments}
We demonstrate the effectiveness of our proposed DMVAE framework on the semi-supervised classification setting, commonly adopted as a benchmark for multimodal generative models \cite{VedantamFH018,NIPS2018_7801}. We also evaluate our model qualitatively by cross-synthesizing images from the label modality. \autoref{sec5.single} summarizes the single-label classification results. In \autoref{sec5.multi}, we further investigate DMVAE on the multi-label classification problem with highly imbalanced labels. \autoref{sec5.abl} evaluates the effectiveness of different model components in an ablation study.

\subsection{Single-Label Classification} \label{sec5.single}
As in MVAE \cite{NIPS2018_7801}, we ground the bi-modal setup by giving one modality as images and another modality as the image class labels. By assuming that only a portion of the input data is paired, we arrive to what is a traditional weakly-supervised learning setting.  We consider MNIST and Fashion MNIST (FMNIST) as the benchmark datasets. The digit identity $\{0,\ldots,9\}$ on MNIST and clothes type $\{$T-shirt/top, Trouser, Pullover, Dress, Coat, Sandal, Shirt, Sneaker, Bag, Ankle boot$\}$ on FMNIST is expected to be the shared information between the image and label modalities. 
While the label modality has only the shared latent space, the image modality also includes the private latent space, generally assumed to model the styles of the digits, such as the stroke thickness, width, tilt, etc. 50,000 examples are used as the training set of both MNIST and FMNIST. Both datasets have the continuous private space of ten dimensions for the image modality and one discrete shared latent variable of ten dimensions to model the ten class factors. The details of the model (encoder/decoder) architectures and the optimization are described in the Supplement.

\begin{table}[t]
\caption{Classification accuracy for MNIST and FMNIST as a function of the paired data fraction (in \% points). Bold denotes the highest accuracy.}
\label{tab:mnist}
\begin{center}
\scriptsize

\setlength\tabcolsep{2.5pt}
\begin{tabular}
{lN{1}{3}N{1}{3}N{1}{3}N{1}{3}N{1}{3}N{1}{3}N{1}{3}N{1}{3}
N{2}{1}
N{1}{3}N{1}{3}N{1}{3}N{1}{3}N{1}{3}N{1}{3}N{1}{3}N{1}{3}
}

\toprule
 Model & \multicolumn{8}{c}{MNIST} & \multicolumn{9}{c}{FMNIST} \\
\midrule
  & {0.1}  & {0.2}  & {0.5}  & {{1}}   & {{5}}    & {10}   & {50}  & {100}
&& {0.1}  & {0.2}  & {0.5}  & {{1}}   & {{5}}    & {10}   & {50}  & {100}
  \\ 
\midrule
AE     & 0.4143 & 0.5429 & 0.6448 & 0.788  & 0.9124 & 0.9269 & 0.9423 & 0.9369 
& & {{-}} & {{-}} & {{-}} & {{-}} & {{-}} & {{-}} & {{-}} & {{-}}
\\
NN     & 0.6618 & 0.6964 & 0.7971 & 0.8499 & 0.9235 & 0.9455 & 0.9806 & 0.9857 
& & \topscore 0.6755 & 0.701  & 0.7654 & 0.7944 & 0.8439 & 0.862  & 0.8998 & 0.9318
\\
LOGREG & 0.6565 & 0.7014 & 0.7907 & 0.8391 & 0.8713 & 0.8665 & 0.9217 & 0.9255 
& & 0.6612 & 0.7005 & 0.7624 & 0.7627 & 0.7802 & 0.8015 & 0.8377 & 0.8412 \\
RBM    & 0.7152 & 0.7496 & 0.8288 & 0.8614 & 0.917  & 0.9257 & 0.9365 & 0.9379 
& & 0.6708 & 0.7214 & 0.7628 & 0.769  & 0.7943 & 0.8021 & 0.8088 & 0.8115 
\\
VAE    & 0.2547 & 0.284  & 0.4026 & 0.6369 & 0.8717 & 0.8989 & 0.9183 & 0.9311 
& & 0.5316 & 0.6502 & 0.7221 & 0.7324 & 0.7697 & 0.7765 & 0.7914 & 0.8311
\\
JMVAE  & 0.2342 & 0.2809 & 0.3386 & 0.6116 & 0.8638 & 0.9051 & 0.9498 & 0.9572 
& & 0.5284 & 0.5737 & 0.6641 & 0.6996 & 0.7937 & 0.8212 & 0.8514 & 0.8828
\\
MVAE  \footnote[3]    
& 0.4409 & 0.5793 & 0.829 & 0.8967 & 0.935 & 0.9481 & 0.9613 & 0.9711
& & 0.4865 & 0.5591 & 0.66 & 0.7725 & 0.8265 &	0.8423 & 0.8692 & 0.882
\\
\midrule
DMVAE  & \topscore 0.7735 & \topscore 0.9166 & \topscore 0.9443 & \topscore 0.9455 & \topscore 0.9573 & \topscore 0.9605 & \topscore 0.9740 & \topscore 0.9810 
& & 0.5406 & \topscore 0.7490 & \topscore 0.7774 & \topscore 0.8335 & \topscore 0.8709 & \topscore 0.8937 & \topscore 0.9280 & \topscore 0.941  \\

\bottomrule                    
\end{tabular}
\end{center}
\end{table}

\begin{figure*}[t]
\centering

\begin{subfigure}[b]{0.57\textwidth} 
{\includegraphics[width=\textwidth]{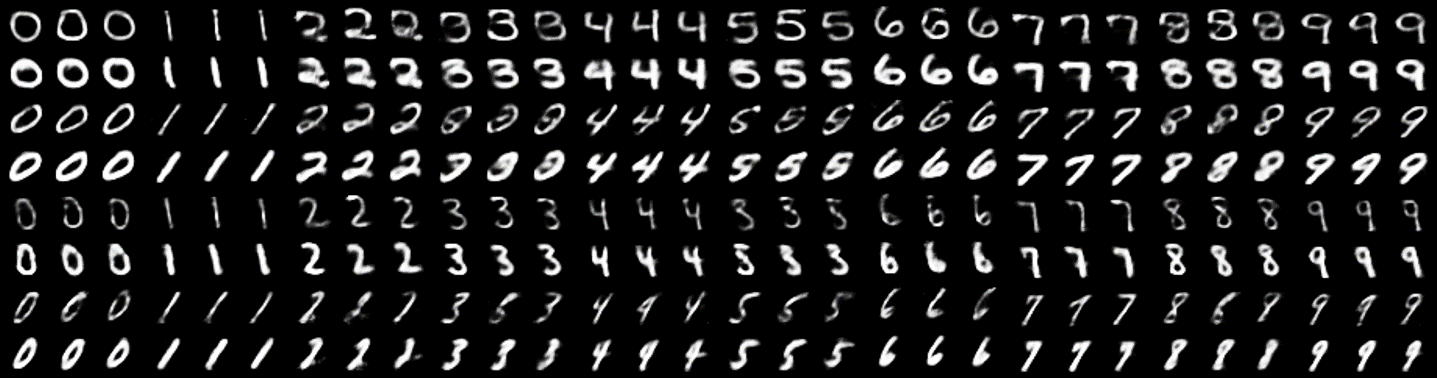} \caption{}\label{fig:mnist_cross} }
\end{subfigure}
\hfill
\begin{subfigure}[b]{0.38\textwidth} 
 {\includegraphics[width=\textwidth]{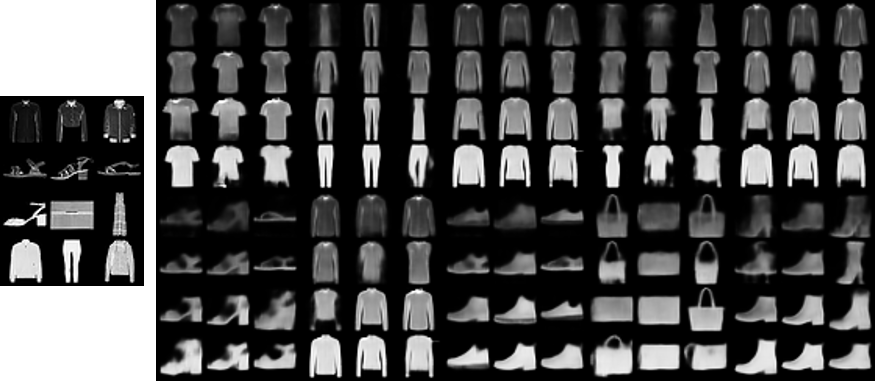} \caption{}\label{fig:fmnist_cross} }
 \end{subfigure}
 
  \caption{Cross-synthesis images from labels in the missing modality case using the model trained with 0.2\% paired data.
  (a) MNIST results where $z_{p,Image} \sim  {\cal N}(0,1)$. (b) FMNIST results  where the left 12 images are the true test images from which $z_{p,Image}$ comes. }

\label{fig:cross}
\end{figure*}

\noindent\textbf{Quantitative Evaluation.} At test time, the label is cross-generated from the test input image by transferring the inferred shared latent factors of the image modality into the "reconstructed" label modality. Thus, the cross-modal inference in our generative model becomes the process of classification. We compare our results against MVAE ~\cite{NIPS2018_7801}\footnote{We use the authors' publicly available code for MVAE to reproduce the baseline results.  Note that we, as well as others, were unable to reproduce the results published in~\cite{NIPS2018_7801}, even after extensive hyperparameter search. We report the best performance found.  See Supplement for more details. }. \autoref{tab:mnist} shows the classification accuracy on MNIST and FMNIST.
On MNIST, even with very few paired data such as 0.1, 0.2, or 0.5 \% which uses 50, 100, 250 labels respectively, we achieve significantly better accuracy than MVAE, suggesting our model is effective in weakly-supervised learning settings. Similarly on FMNIST, DMVAE outperforms MVAE over all paired data fractions. 
These results underline the ability of DMVAE to disentangle latent factors and distil them into the shared factor representing the digit label and the private factors surmising the image style, even with only few labeled examples.
In contrast to MNIST, FMNIST cannot be classified with more than 0.9 accuracy when 0.2\% data are paired. We further analyze this in the Supplement.

 


\noindent\textbf{Qualitative Evaluation.} \autoref{fig:cross} shows the cross-synthesized images conditioned on a target label. For the image modality to be inferred by the generative model, both its private and shared latent factors are necessary. The shared factor is determined by the conditioning label modality. The private image factor can be either sampled from the prior $p(z_{Image})$ or formed by arbitrary "style" transfer conditioning, $q(z_{Image}|x_{Image,style})$ (e.g., asking the model to create an image of digit "2" in the style of the conditioning image of digit "9").  We demonstrate the results of both  setting; the former on MNIST and the later on FMNIST. Additional results can be found in the Supplement.
In \autoref{fig:mnist_cross}, each set in a 8 $\times$ 3 block of images has the same conditioning label. For the style of each image in the block, we first sample the private latent factor  from the Gaussian prior distribution, and replace the three latent values with "extreme" style -2 or +2 to assess whether the visibly distinct style (width, slant, thickness) is kept. From top to bottom, those values are (-,-,-), (-,-,+), (-,+,-), (-,+,+), (+,-,-), (+,-,+), (+,+,-), (+,+,+). Thus, the first row should depict narrow, not-slanted, and thin style, while the last row has wide, slanted, and thick style. Each 8 $\times$ 3 block is synthesized reflecting the conditioning digit class label. 
In \autoref{fig:fmnist_cross}, we illustrate the results of FMNIST using the test image for style transferring. Each set in a 4 $\times$ 3 block of images has the same conditioning label, indicating the ability of DMVAE to correctly disentangle the two modalities.  We see that the style, brightness, is correctly transferred.


\begin{table*}[t]

\caption{(a) Classification F1-score and accuracy for CelebA according to the paired data fraction. The last column is quantitative evaluation on cross-synthesized images of the five most balanced attributes. (b) Accuracy for MNIST under the model trained with 0.2\% paired data. D($C/D$,i) means DMVAE with continuous/discrete shared latent space and TC weight of image modality = i.}
\begin{center}
\scriptsize

\setlength\tabcolsep{2.5pt}
\begin{subtable}[t]{.48\textwidth}
\begin{tabular}
{llN{1}{3}N{1}{3}N{1}{3}N{1}{3}N{1}{3} N{2}{1} N{1}{3}}
\toprule
   &   & {1\%}    & {10\%}   & {50\%}   & {100\%} && {cross-syn} \\ 
\midrule
\multirow{2}{*}{F1-score} & {MVAE}  &  0.4438 &	0.5130 & 0.5289 & 0.5140 
&& 0.6681 \\
&{DMVAE} & \topscore 0.5219 & \topscore 0.6269 & \topscore 0.6873 & \topscore  0.7036 
&& \topscore 0.7264
\\ \midrule
\multirow{2}{*}{Accuracy} & {MVAE}  & \topscore 0.8625 & 0.8791 & 0.8904 & 0.8916 
&& 0.5096 \\
&{DMVAE} & 0.8615 & \topscore 0.8875 & \topscore 0.9020 & \topscore 0.9072
&& \topscore 0.609 \\
\bottomrule
\end{tabular}
\caption{}\label{tab:celeba_f1}
\end{subtable}
\hfill
\begin{subtable}[t]{.48\textwidth}
\setlength\tabcolsep{2.5pt}
\begin{tabular}
{lN{1}{3}N{1}{3}N{1}{3}N{1}{3}N{1}{3}}
\toprule
  & {MVAE} & {D(c,1)} & {D(c,3)} & {D(d,1)} & {D(d,3)} \\
\midrule
Accuracy & 0.6254 & 0.6384	& 0.6946 &  0.8032 & \topscore 0.9166 \\
\bottomrule
\end{tabular}
\caption{}\label{tab:ablation}
\end{subtable}

\end{center}
\vspace{-1.5em}
\end{table*}

\subsection{Multi-Label Classification}  \label{sec5.multi}
We further examine an application of DMVAE in the weakly-supervised learning for the multi-label classification problem on CelebA \cite{liu2015faceattributes} dataset. CelebA consists of images of celebrities with 40 attributes, such as attractiveness, age group, gender, etc. As in MVAE \cite{NIPS2018_7801}, we assess the performance of predicting 18 visually distinct attributes. Since most of the attributes are highly imbalanced \footnote{The ratio of positive labels for the 18 attributes in training set is provided in the Supplement.}, we use both the F1-score and the classification accuracy as performance indexes. We compare our results with MVAE by reproducing the bi-modality setting with the authors' publicly available code. We set 100 latent dimensions for the continuous private space of the image modality and 18 latent dimensions for the discrete shared space for both image and the attribute modality. Each shared latent factor consists of a binary discrete variable to represent the activation of individual attributes.

\noindent\textbf{Quantitative Evaluation.} \autoref{tab:celeba_f1} reports the classification accuracy and F1-scores. Each attribute is evaluated, with performance averaged over all 18 attributes. The image modality and the label modality again have the common shared space, with the aim of representing the 18 attributes. Only the image modality has the private space, which should represent the remaining 22 attributes and and image-specific information, such as the background etc. We predict the label of each attribute by cross-generating it from the shared latent factors, inferred from the image modality. For both our model and the baseline, the accuracy increases as the paired data fraction grows. In each case, we beat the baseline. The F1-score of MVAE reduces with 100 \% paired data, while DMVAE's performance improves. This implies DMVAE is robust to imbalanced data, in contrast to MVAE.

\begin{figure*}[t]
\begin{center}
\begin{subfigure}[b]{0.185\textwidth}
  {\includegraphics[width=\textwidth]{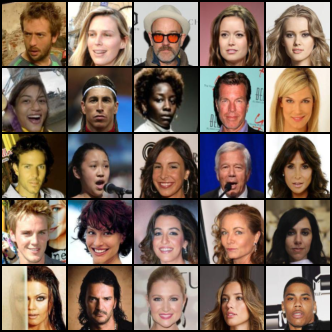} \caption{}\label{fig:celeb_gt} }
\end{subfigure}
\hfill
\begin{subfigure}[b]{0.185\textwidth}
  {\includegraphics[width=\textwidth]{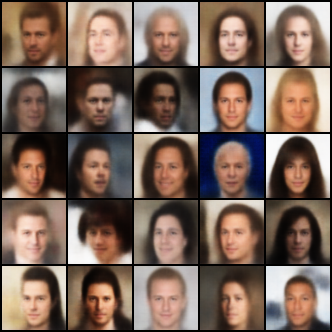} \caption{} }
\end{subfigure}
\hfill
\begin{subfigure}[b]{0.185\textwidth}
   {\includegraphics[width=\textwidth]{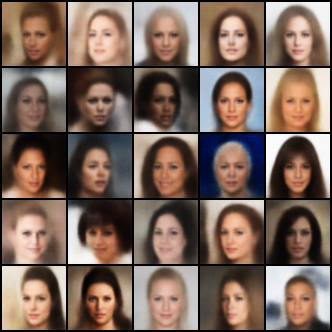} \caption{} }
\end{subfigure}
\hfill
\begin{subfigure}[b]{0.185\textwidth}
    {\includegraphics[width=\textwidth]{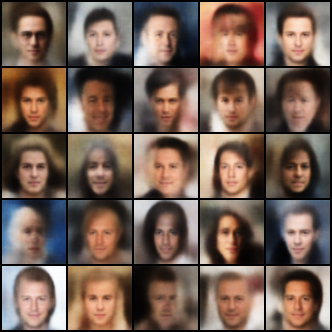} \caption{}}
\end{subfigure}
\hfill
\begin{subfigure}[b]{0.185\textwidth}
  {\includegraphics[width=\textwidth]{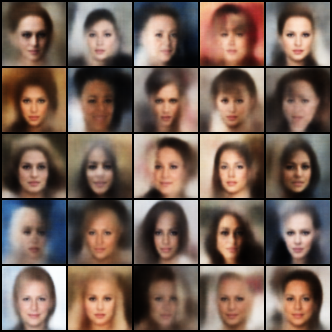} \caption{}}
\end{subfigure}
\end{center}
  \caption{Results under the model trained with 1\% paired data of training set.  (a): Ground-truth images from test set. (b) and (c): Cross-synthesized images with  `Male' and `Heavy Makeup' attribute respectively where $z_{p_2}$ is from (a). (d) and (e): Cross-synthesized images with `Male' and `Heavy Makeup' attribute respectively where $z_{p_2}$ is from Gaussian prior distribution. }
\label{fig:celeba_cross}
\end{figure*}


\noindent\textbf{Qualitative Evaluation.} We cross-synthesize an image conditioned on the target attribute by transferring the latent factor of the attribute modality into the image modality through the shared space. \autoref{fig:celeba_cross} shows the results conditioned on `Male' or `Heavy Makeup' attribute. 
We inspect these results in a quantitative manner by predicting the conditioned target attribute reversely from the synthesized images. For this,  separate CNN classifiers are trained for DMVAE and MVAE. We evaluate the averaged accuracy and F1-score of the five most balanced attributes - heavy makeup, male, mouth slightly open, smiling, and wavy hair. As in the last column of \autoref{tab:celeba_f1}, DMVAE achieves better performance than MVAE. 
This suggests DMVAE can effectively gather and transfer only the required attribute information, disentangling it from the irrelevant factors from the private space. In the Supplement, we investigate the importance of each factor to predict the target attribute more explicitly by applying an SVM with a L1 regularizer onto the features of images with the target attribute.


\subsection{Ablation Study} \label{sec5.abl}
We examine the effectiveness of each component of DMVAE on MNIST in \autoref{tab:ablation}. All DMVAE settings provide better performance than MVAE. Within DMVAE, discrete factors improve prediction as they are better aligned with the categorical nature of attributes. We also assess the importance of enforcing disentanglement by varying the weights $\beta$ of the TC term. With higher TC weights, the shared latents achieve better disentanglement than the private ones,  increasing the accuracy.

\section{Conclusion}
In this paper, we introduce a novel multi-modal VAE model with separated private and shared spaces. We verify that having a private space per modality as well as the common shared space can significantly impact the representational performance of multimodal VAE models. We also demonstrate that disentanglement is critical for separation of factors across the two sets of spaces. Finally, the presence of continuous and discrete latent representations is important in instances where some of the attributes naturally exhibit categorical properties.

\section*{Broader Impact}

\begin{enumerate}
    \item \textbf{Who may benefit from this research?} For any individuals, practitioners, organizations, and groups who aim to explore multimodal data, this research can be a useful tool that provides interpretable models for cross-modal data linking and prediction within and across modalities. 
    \item \textbf{Who may be put at disadvantage from this research?} Not particularly applicable. 
    \item \textbf{What are the consequences of failure of the system?} Any failure of the system that implements our algorithm would not do any serious harm since the failure can be easily detectable at the validation stage, in which case alternative strategies or internal decisions might be looked for.  
    \item \textbf{Whether the task/method leverages biases in the data?} Our method does not leverage biases in the data. 
\end{enumerate}

\newpage

\bibliographystyle{bib_style}
\bibliography{nips}

\newpage

\textbf{\LARGE Appendix: Private-Shared Disentangled Multimodal VAE for Learning of Hybrid Latent Representations
}




\appendix

\noindent \\This supplement consists of the following materials: 

\begin{itemize}
\item Additional details related to concepts in the main paper in \autoref{sec:supplementary}.
\item Details of network architectures in \autoref{sec:netarch}.
\item Optimization details in \autoref{sec:optim}.
\item Additional experimental results in \autoref{sec:experiments}.
\end{itemize}

\section{Concept Motivation and Additional Details}\label{sec:supplementary}

\subsection{Private \& Shared Spaces}

\begin{figure}[h]
\centering
 \begin{subfigure}[b]{0.48\textwidth}
 {\includegraphics[width=\textwidth]{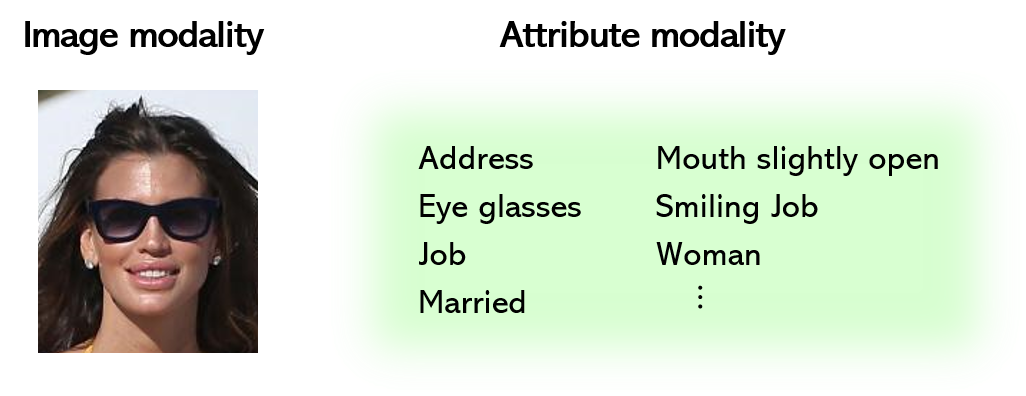} \caption{Bimodal data}\label{fig:two_modal} }
 \end{subfigure}
 \hfill
\begin{subfigure}[b]{0.48\textwidth} 
{\includegraphics[width=\textwidth]{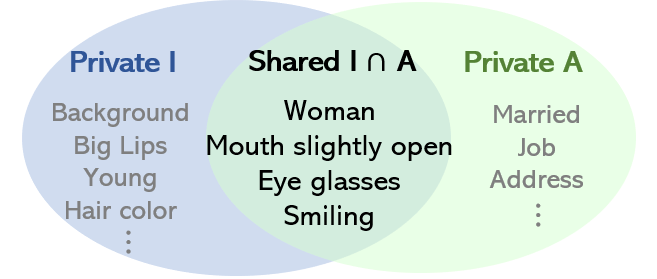} \caption{Private and shared spaces used by DMVAE}\label{fig:two_space}}
\end{subfigure}
  \caption{(a) Example of bimodal data, where one modality, $\texttt{I}$, is an image of a person and the other, $\texttt{A}$, represents general personal attributes. (b)  Only some of the factors, here aligned with attributes for simplicity, are shared by both modalities in $\texttt{Shared I} \cap \texttt{A}$. Other factors are private to individual modalities, grouped in separate $\texttt{Private}$ spaces.  By definition, the three spaces are \textbf{disentangled} from each other.}
\label{fig:two}
\end{figure}

We pictorially highlight the concept of private and shared spaces modeling, using the motivating example of the human face images as one modality and personal attributes as another.   In \autoref{fig:two_modal}, 
an image of a smiling woman with eye glasses is augmented with labels describing both visual and non-visual attributes of that individual.  Visual attributes, such as the woman displaying open mouth, smiling, and having eyeglasses, have their counterparts in the image domain.  However, other attributes such as the woman's marital status, job, etc., cannot be deduced from the image view.  Similarly, many image attributes are not captured in the accompanied labels.  Therefore, accurate modeling of the underlying data representation has to consider both the \textbf{private} aspects of individual modalities as well as what those modalities \textbf{share}, as illustrated in \autoref{fig:two_space}.

\subsection{Concrete Distribution for Shared Space Modeling}
\begin{figure}[tbhp]
\centering
 \begin{subfigure}[b]{0.48\textwidth}
 {\includegraphics[width=\textwidth]{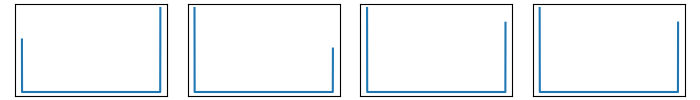} \caption{$T$ = 0.01}\label{fig:poe1} }
 \end{subfigure}
 \begin{subfigure}[b]{0.48\textwidth}
 {\includegraphics[width=\textwidth]{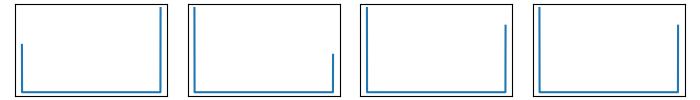} \caption{$T$ = 0.66}\label{fig:poe2} }
 \end{subfigure}
\begin{subfigure}[b]{0.48\textwidth}
 {\includegraphics[width=\textwidth]{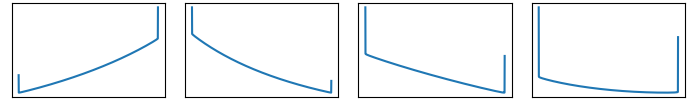} \caption{$T$ = 0.999}\label{fig:poe3} }
 \end{subfigure}
 \begin{subfigure}[b]{0.48\textwidth}
 {\includegraphics[width=\textwidth]{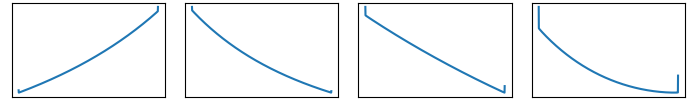} \caption{$T$ = 0.9999}\label{fig:poe4} }
 \end{subfigure}
 \begin{subfigure}[b]{0.48\textwidth}
 {\includegraphics[width=\textwidth]{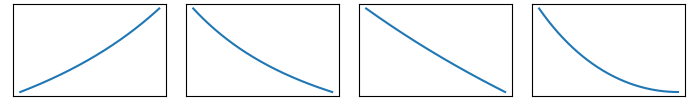} \caption{$T$ = 1}\label{fig:poe5} }
 \end{subfigure}
 \begin{subfigure}[b]{0.48\textwidth}
 {\includegraphics[width=\textwidth]{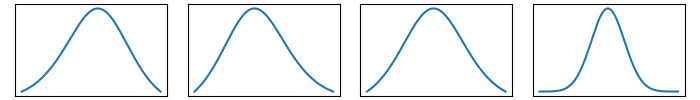} \caption{$T$ = 2}\label{fig:poe6} }
 \end{subfigure}
  \caption{PDF of a random variable $z = [z_1, z_2] \in \mathbb{R}^2$ where $z_i \in [0,1]$. For each $T$, the first three plots illustrate z $\sim$ Concrete($\pi_i, T$) with $\pi_1 = [4,3], \pi_2 = [2,3], \pi_3 = \pi_1 \times \pi_2 = [8,9]$, in order. The last plot is the PDF of the normalized product-of-expert of  $z \sim$ Concrete($\pi_1, T$) and  $z \sim$ Concrete($\pi_2, T$). X-axis is from $[0,1]$ to $[1,0]$. As $T$ increases, the concrete distribution deviates from the categorical distribution and shows states on non-vertices. We use $T = 0.66$ in our experiments.   }
\label{fig:poe}
\end{figure}

In the main paper, to approximate equation (8), we require the property of the concrete distribution on the simplex. 
The argmax computation of the categorical distribution has the property that it returns states on the vertices of the simplex $\Delta^{n-1}$. Because the concrete distribution approximates the categorical distribution arbitrarily closely as $T \rightarrow 0$, it satisfies the desired property. \autoref{fig:poe} shows an example of two discrete densities $\pi_1 = [4,3], \pi_2 = [2,3]$ that give rise to the PoE density $\pi_3 = \pi_1 \times \pi_2 = [8,9]$.  Both are approximated using the concrete approximation at different temperatures $T$. Their PoE is a product of concrete densities, with the approximation illustrated in the fourth panel for each $T$.   Note that the concrete PoE approximation closely follows the true discrete density PoE.

\section{Neural Network Architecture}\label{sec:netarch}
We describe our model architecture in \autoref{tab:mnist_arch}, \autoref{tab:fmnist_arch}, and \autoref{tab:celeba_arch} for MNIST, FMNIST, and CelebA respectively. $Z_{p,I}$ and $Z_{s}$ indicate the latent dimension of the private space of image modality and the latent dimension of the shared space, respectively.

\begin{table}[tbhp]
\caption{MNIST network.}
\label{tab:mnist_arch}
\begin{center}
\scalebox{0.95}{\begin{tabular}{p{3.5cm}p{3.7cm}p{2.5cm}p{2.7cm}}
\toprule
\multicolumn{2}{c}{Image}                                                                        & \multicolumn{2}{c}{Label} \\                     \midrule     
\multicolumn{1}{c}{Encoder}                                        & \multicolumn{1}{c}{Decoder} & \multicolumn{1}{c}{Encoder} & \multicolumn{1}{c}{Decoder} \\
\hline
Input: Image (1$\times$28$\times$28) & Input: Latents ($Z_{p,I} + Z_s$)       
& Input: Label (10)           & Input: Latents ($Z_{s}$)               \\
Linear 784 $\times$  256                                                    & Linear ($Z_{p,I} + Z_s$) $\times$  256           & Linear 10 $\times$  256              & Linear $Z_s$ $\times$  256           \\
ReLU                                                               & ReLU                        & ReLU                        & ReLU                        \\
Linear 256 $\times$ ($Z_{p,I} + Z_s$) & Linear 256 $\times$  784             & Linear 256 $\times$ $Z_s$            & Linear 256 $\times$  10              \\
                                                                   & Sigmoid                     &                             & Sigmoid            \\
\bottomrule
\end{tabular}}
\end{center}
\end{table}

\begin{table}[]
\caption{FMNIST network.}
\label{tab:fmnist_arch}
\begin{center}
\scalebox{0.95}{\begin{tabular}{p{4cm}p{4cm}p{2.3cm}p{2.3cm}}
\toprule
\multicolumn{2}{c}{Image}                                                                        & \multicolumn{2}{c}{Label} \\                     \midrule     
\multicolumn{1}{c}{Encoder}                                        & \multicolumn{1}{c}{Decoder} & \multicolumn{1}{c}{Encoder} & \multicolumn{1}{c}{Decoder} \\
\hline

Input: Image (1$\times$28$\times$28)           & Input: Latents ($Z_{p,I} + Z_s$)         & Input: Label (10)           & Input: Latent ($Z_s$)       \\
32 Conv 3 $\times$ 3, stride 1, padding 1  & Linear ($Z_{p,I}+Z_s$) $\times$ 512, ReLU   & Linear 10 $\times$ 256      & Linear $Z_s$ $\times$ 256   \\
ReLU                      & Linear 512 $\times$  (64$\times$7$\times$7)                   & ReLU                        & ReLU                        \\
32 Conv 3 $\times$ 3, stride 1, padding 1  & ReLU, Upsample  & Linear 256 $\times$ 10      \\
ReLU, MaxPool, Dropout 0.25 &  64 Conv 3 $\times$ 3, stride 1, padding 1 
& Linear 256 $\times$ $Z_s$  & Sigmoid                     \\
64 Conv 3 $\times$ 4, stride 1, padding 1 & ReLU  &                             &                             \\
ReLU                            & 32 Conv 3 $\times$ 3, stride 1, padding 1 &                             &                             \\
64 Conv 3 $\times$ 4, stride 1, padding 1 & ReLU, Upsample  &                             &                             \\
ReLU, MaxPool, Dropout 0.25                           & 32 Conv 3 $\times$ 3, stride 1, padding 1                         &                             &                             \\
Linear (64$\times$7$\times$7) $\times$ 512     & ReLU   &                             &                             \\
ReLU, Dropout 0.5                           & 1 Conv 3 $\times$ 3, stride 1, padding 1      &                             &                             \\
Linear 512 $\times$ ($Z_{p,I} + Z_s$)    & Sigmoid     \\
\bottomrule
\end{tabular}}
\end{center}
\end{table}

\begin{table}[]
\caption{CelebA network.}
\label{tab:celeba_arch}
\begin{center}
\scalebox{0.95}{\begin{tabular}{p{3.5cm}p{4.6cm}p{2.3cm}p{2.3cm}}
\toprule
\multicolumn{2}{c}{Image}                                                                        & \multicolumn{2}{c}{Label} \\                     \midrule     
\multicolumn{1}{c}{Encoder}                                        & \multicolumn{1}{c}{Decoder} & \multicolumn{1}{c}{Encoder} & \multicolumn{1}{c}{Decoder} \\
\hline
Input: Image (3$\times$64$\times$64)          & Input: Latents ($Z_{p,I} + Z_s$)          & Input: Label (18)           & Input: Latent ($Z_s$)       \\
32 Conv 4 $\times$ 4, stride 2  & Linear ($Z_{p,I} + Z_s$) $\times$ (256$\times$5$\times$5)   & Linear 18 $\times$ 256      & Linear $Z_s$ $\times$ 256   \\
ReLU                            & ReLU                            & ReLU                        & ReLU                        \\
64 Conv 4 $\times$ 4, stride 2  & 256 Conv 4 $\times$ 4, stride 2 & Linear 256 $\times$ $Z_s$   & Linear 256 $\times$ 18      \\
ReLU                            & ReLU                            &                             & Sigmoid                     \\
128 Conv 4 $\times$ 4, stride 2 & 128 Conv 4 $\times$ 4, stride 2 &                             &                             \\
ReLU                            & ReLU                            &                             &                             \\
256 Conv 4 $\times$ 4, stride 2 & 64 Conv 4 $\times$ 4, stride 2  &                             &                             \\
ReLU                            & ReLU                            &                             &                             \\
Linear (256$\times$5$\times$5) $\times$ 512     & 32 Conv 4 $\times$ 4, stride 2  &                             &                             \\
ReLU                            & Sigmoid                         &                             &                             \\
Dropout 0.1                     &                                 &                             &                             \\
Linear 512 $\times$ ($Z_{p,I} + Z_s$)    &                                 &                             &  \\
\bottomrule
\end{tabular}}
\end{center}
\end{table}

\section{Training Details}\label{sec:optim}
We use Adam optimizer and 100 batch size for all datasets except for the 0.1\% and 0.5\% paired data fractions of MNIST and FMNIST since they use 50 and 250 labeled samples respectively which requires size 50 batch. 
Because this is a weakly-supervised learning setting, in order to assess the uncertainty stemming from random selection of labels, we train the model with three different seeds, 0,1, and 2. 
Other details, specific to each dataset, are summarized in \autoref{tab:hyper}.

\begin{table}[h]
\caption{Optimization Details.}
\label{tab:hyper}
\begin{center}
\begin{tabular}{p{3.4cm}p{2.6cm}p{2.6cm}p{2.6cm}}
\toprule
\multicolumn{1}{c}{Hyper-parameter}                 & \multicolumn{1}{c}{MNIST} & \multicolumn{1}{c}{FMNIST} & \multicolumn{1}{c}{CelebA} \\ \midrule
Private latent variables             & 10 continuous             & 10 continuous            & 100 continuous             \\
Shared latent variables              & 1 10-D discrete           & 1 10-D discrete          & 18 binary discrete         \\
Learning rate                        & $1e^{-3}$                 & $1e^{-3}$                & $1e^{-4}$                  \\
Epoch                                & 500                       & 400                      & 150                        \\
\bottomrule
\end{tabular}
\end{center}
\end{table}

\begin{figure*}[h]
\centering
\begin{subfigure}[b]{0.3\textwidth}
 {\begin{tikzpicture}[
varNode/.style={%
    circle,
    draw,
    thick,
    inner sep=3pt, 
    minimum size=22pt
  }
]

\node[varNode,fill=gray!30] (x1) at (1,1) {$x_1$};
\node[varNode,fill=gray!30] (x2) at (3,1) {$x_2$};

\node[varNode,fill=white] (z1p) at (0,0) {$z_{p_1}$};
\node[varNode,fill=white] (zs) at (2,0) {$z_s$};

\node[varNode,fill=gray!30] (x1t) at (1,-1) {$\Tilde{x}_1$};
\node[varNode,fill=gray!30] (x2t) at (3,-1) {$\Tilde{x}_2$};

\draw[->,thick] (x1) -- (z1p) ;
\draw[->,thick] (x2) -- (zs) ;
\draw[->,thick] (z1p) -- (x1t) ;
\draw[->,thick] (zs) -- (x2t) ;
\draw[->,thick] (zs) -- (x1t) ;

\draw[green,line width=5pt,opacity=0.4,] (x1.center) -- (z1p.center) ;
\draw[->,green,line width=5pt,opacity=0.4,] (z1p.center) -- (x1t) ;

\draw[yellow,line width=5pt,opacity=0.4,] (x2.center) -- (zs.center) ;
\draw[->,yellow,line width=5pt,opacity=0.4,] (zs.center) -- (x1t) ;

\end{tikzpicture} \caption{Cross-reconstruction with conditioning}\label{fig:cr_infer_cond} }
 \end{subfigure}
\hspace{0.5cm}
\begin{subfigure}[b]{0.3\textwidth} 
{\pgfmathdeclarefunction{gauss}{2}{\pgfmathparse{1/(#2*sqrt(2*pi))*exp(-((x-#1)^2)/(2*#2^2))}}

\begin{tikzpicture}[
varNode/.style={%
    circle,
    draw,
    thick,
    inner sep=3pt, 
    minimum size=22pt
  }
]

\node[varNode,fill=gray!30] (x2) at (3,1) {$x_2$};

\node[varNode,fill=white] (z1p) at (0,0) {$z_{p_1}$};
\node[varNode,fill=white] (zs) at (2,0) {$z_s$};

\node[varNode,fill=gray!30] (x1t) at (1,-1) {$\Tilde{x}_1$};
\node[varNode,fill=gray!30] (x2t) at (3,-1) {$\Tilde{x}_2$};

\draw[->,thick] (x2) -- (zs) ;
\draw[->,thick] (z1p) -- (x1t) ;
\draw[->,thick] (zs) -- (x2t) ;
\draw[->,thick] (zs) -- (x1t) ;

\draw[->,green,line width=5pt,opacity=0.4,] (z1p.center) -- (x1t) ;

\draw[yellow,line width=5pt,opacity=0.4,] (x2.center) -- (zs.center) ;
\draw[->,yellow,line width=5pt,opacity=0.4,] (zs.center) -- (x1t) ;

\begin{axis}[xshift=-.55cm,yshift=.4cm,width=2.5cm, axis line style={draw=none}, tick style={draw=none}, xticklabels={},yticklabels={}]
\addplot [draw=black, domain=-3:3] {gauss(0,1)};
\end{axis}

\end{tikzpicture} \caption{Cross-reconstruction with missing modality}\label{fig:cr_infer_missing}}
\end{subfigure}
\hspace{0.5cm}
\begin{subfigure}[b]{0.3\textwidth} 
 {\begin{tikzpicture}[
varNode/.style={%
    circle,
    draw,
    thick,
    inner sep=3pt, 
    minimum size=22pt
  }
]

\node[varNode,fill=gray!30] (x1) at (1,1) {$x_1$};

\node[varNode,fill=white] (z1p) at (0,0) {$z_{p_1}$};
\node[varNode,fill=white] (zs) at (2,0) {$z_s$};

\node[varNode,fill=gray!30] (x1t) at (1,-1) {$\Tilde{x}_1$};

\draw[->,thick] (x1) -- (z1p) ;
\draw[->,thick] (z1p) -- (x1t) ;
\draw[->,thick] (zs) -- (x1t) ;

\draw[green,line width=5pt,opacity=0.4,] (x1.center) -- (z1p.center) ;
\draw[->,green,line width=5pt,opacity=0.4,] (z1p.center) -- (x1t) ;

\draw[->,green,line width=5pt,opacity=0.4,] (zs.center) -- (x1t) ;

\draw[->,thick] (x1) -- (zs) ;
\draw[green,line width=5pt,opacity=0.4,] (x1.center) -- (zs.center) ;


\end{tikzpicture}\centering\caption{Reconstruction}\label{fig:cr_infer_complete}}
\end{subfigure}
  \caption{Three instances of reconstruction of an image and a label using the DMVAE model. \textbf{cross-reconstruction with conditioning (a)}:  both modalities $x_1,x_2$ are given; however, $z_{s}$ is assumed to be inferred only from $x_2$, unlike the complete model. This corresponds to the case of "style transfer", where $z_{p_1}$ (the style of $x_1$) is "injected" into the content $z_s$ of $x_2$.  \textbf{cross-reconstruction with missing modality (b)}: depicts the reconstruction where the "style" of $x_1$ is not know, hence, $z_{p_i}$ is sampled from its prior $p(z_p)$, the standard normal distribution.   \textbf{Reconstruction (c)}: in this instance, we conduct traditional $x_1$ to $\Tilde{x}_1$ reconstruction within a single modality.
}
\label{fig:cr_infer_graph}
\end{figure*}
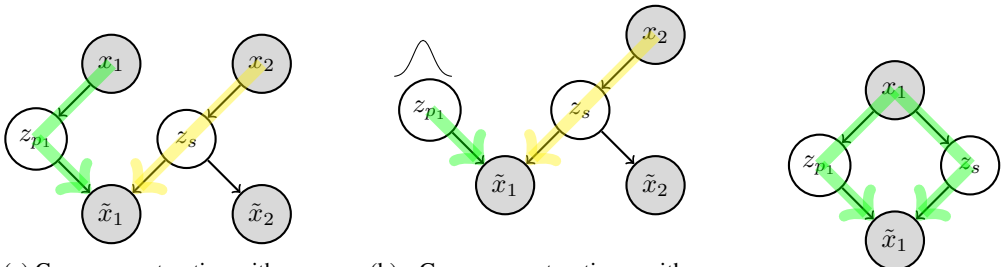

\section{Additional Experimental Results}\label{sec:experiments}

In this section we present additional experimental results on MNIST, FMNIST and CelebA. \autoref{sec3.qual} focuses on studying the quality of cross-synthesized images from labels using DMVAE. We also show additional CelebA cross-synthesis results for different attributes, beyond those in the main paper. \autoref{sec3.sepa} studies the separation of latent spaces (private and shared) using 2-D embeddings, additionally aiming to demonstrate the gains of DMVAE in association of the shared space with the image labels/attributes.  Finally, in \autoref{sec:discrepancy} we highlight the differences between the MVAE results reported in \cite{NIPS2018_7801}  and those which we were able to reproduce for results in Tab. 1 in the main paper.

\subsection{Additional Qualitative Results} \label{sec3.qual}
\begin{figure}[htbp]
\begin{center}
{\includegraphics[width=0.7\textwidth]{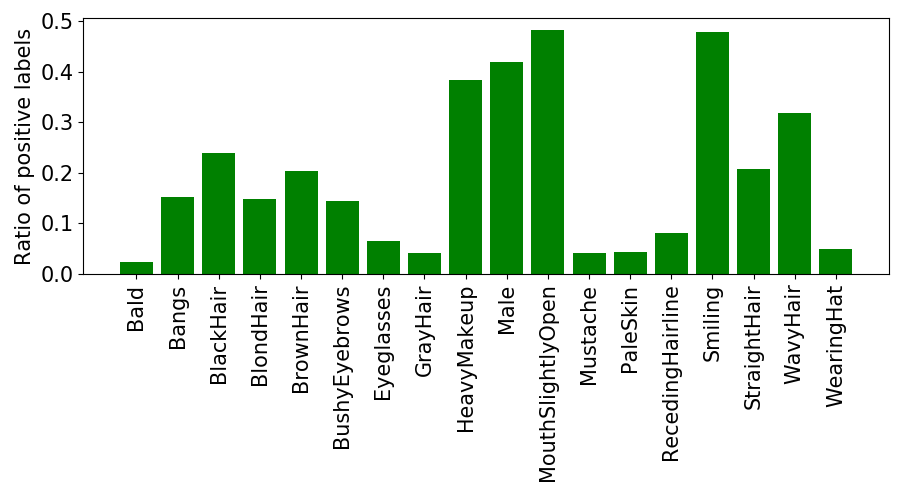}}
\end{center}
 \caption{Ratio of positive labels for 18 visually distinct attributes in training set of CelebA. Ratios close to 0.5 indicate well-balanced attributes. }
\label{fig:celeba_stat}
\end{figure}

We conduct further qualitative evaluation on MNIST, FMNIST, and CelebA. 
For completeness, \autoref{fig:celeba_stat} shows the statistics (ratio of positive labels for the eighteen attributes) in CelebA training set, which guided our selection of key attributes in some experiments.


In the main paper, Fig. 8, we cross-synthesize CelebA images from `Male' or `Heavy Makeup' attribute-activated label assuming the reconstruction inference settings (Sec. 4.3, Main paper), where the missing input image modality was replaced by sampling from $z_{p,image} \sim {\cal N}(0,1)$ or transferring new styles from other test images.  
We also, in Fig. 2 of the main paper, synthesized MNIST images of digits of all classes from latent values with "extreme" style after sampling $z_{p,image} \sim {\cal N}(0,1)$ were provided to assess whether the visibly distinct style (width, slant, thickness) is kept. For FMNIST, we check the style of test images, brightness, was transferred with each class. 

Here, we consider a more complete set of reconstruction experiments.  Specifically, we consider the three reconstruction instances illustrated in \autoref{fig:cr_infer_graph}.  The first instance, \autoref{fig:cr_infer_cond} corresponds to traditional "style transfer" experiments, where the "style" (private space) of $x_1$ is used to map onto the "content" determined by $x_2$.  This can mean that $x_1$ could be an image of digit '1', while $x_2$ is the class label '2'.  The task would be to create a synthetic image of digit '2' in the style of digit '1'.  The second instance, in \autoref{fig:cr_infer_missing}, is that where there is no conditioning modality $x_1$, hence the "style" of $x_1$ is sampled from the prior.  This corresponds to the task of synthesizing any image of class $x_2$. The third instance, in \autoref{fig:cr_infer_complete}, corresponds to the instance of traditional reconstruction of a data point, where both style and digit information comes from $x_1$.

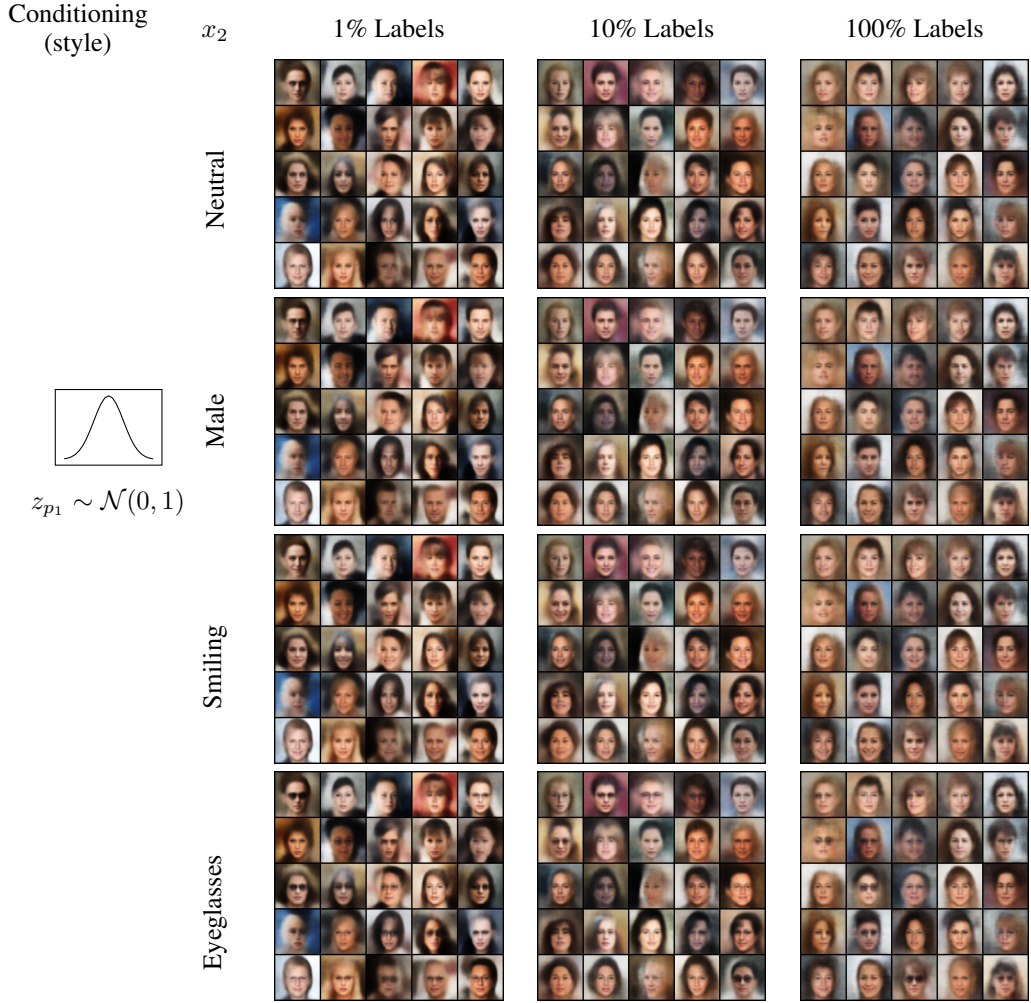
\begin{figure}[tbhp]
\centering
\begin{tabular}{>{\centering\arraybackslash}m{0.15\linewidth}>{\centering\arraybackslash}m{0.05\linewidth}>{\centering\arraybackslash}m{0.22\linewidth}>{\centering\arraybackslash}m{0.22\linewidth}>{\centering\arraybackslash}m{0.22\linewidth}>{\centering\arraybackslash}m{0.22\linewidth}}
Conditioning (style) & $x_2$ & 1\% Labels & 10\% Labels & 100\% Labels \\

&\rotatebox[origin=c]{90}{Neutral}
&{\includegraphics[width=0.99\linewidth]{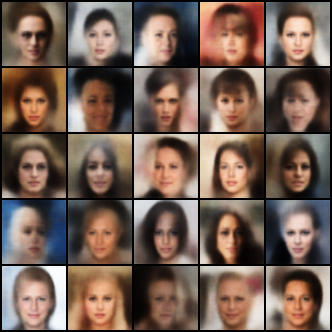}}
&{\includegraphics[width=0.99\linewidth]{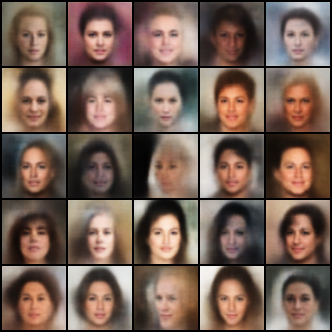}}
&{\includegraphics[width=0.99\linewidth]{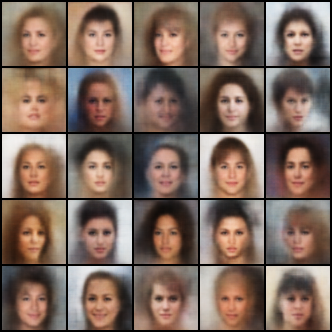}}\\ 

\pgfmathdeclarefunction{gauss}{2}{\pgfmathparse{1/(#2*sqrt(2*pi))*exp(-((x-#1)^2)/(2*#2^2))}}

\multirow{4}{*}{
\begin{tabular}{c}
\begin{tikzpicture}
\begin{axis}[width=3cm, tick style={draw=none}, xticklabels={},yticklabels={}]
\addplot [draw=black, domain=-3:3] {gauss(0,1)};
\end{axis}
\node[] at (.7,-.5) {$z_{p_1} \sim {\cal N}(0,1)$};
\end{tikzpicture}
\end{tabular}
}

 & 
  \rotatebox[origin=c]{90}{Male} 
&{\includegraphics[width=0.99\linewidth]{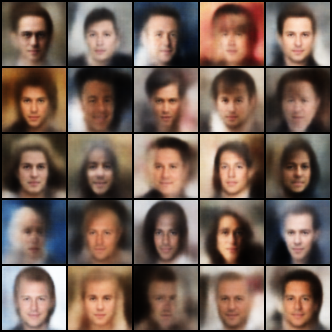}}
&{\includegraphics[width=0.99\linewidth]{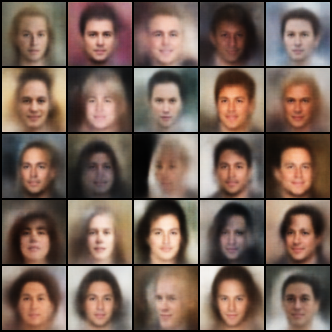}}
&{\includegraphics[width=0.99\linewidth]{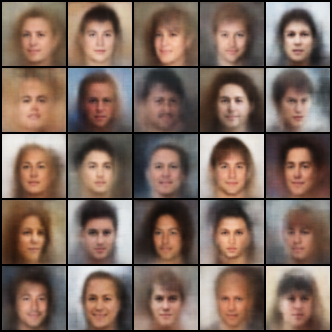}}\\

  & 
\rotatebox[origin=c]{90}{Smiling}  &
{\includegraphics[width=0.99\linewidth]{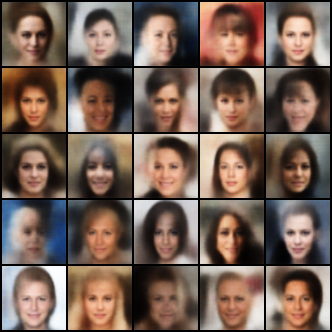}}&
{\includegraphics[width=0.99\linewidth]{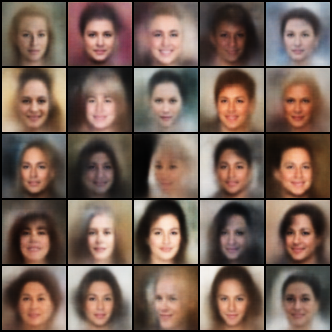}}&
{\includegraphics[width=0.99\linewidth]{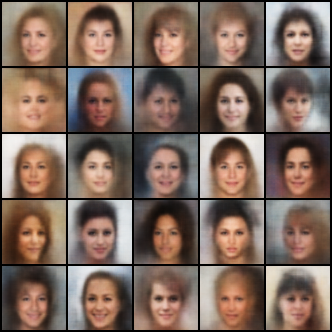}}\\ 

  & 
\rotatebox[origin=c]{90}{Eyeglasses}  &
{\includegraphics[width=0.99\linewidth]{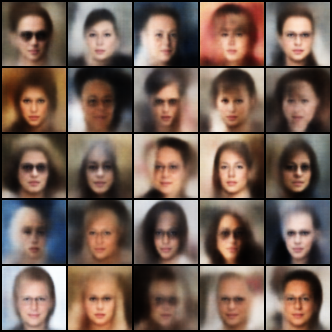}}&
{\includegraphics[width=0.99\linewidth]{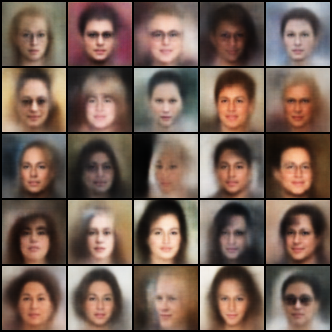}}&
{\includegraphics[width=0.99\linewidth]{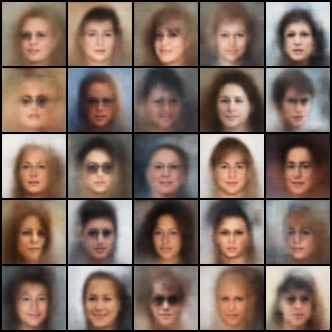}}\\ 
\end{tabular}
    \caption{Cross-synthesized images from attributes, where the image private latent is $z_{p,Image} \sim {\cal N}(0,1)$. Rows of images correspond to no attributes activated (i.e., `Neutral'), `Male', `Smiling', and `Eyeglasses', as depicted.  The columns correspond to DMVAE models trained on data with $1\%$, $10\%$, and $100\%$ labels.  Note that, as expected, the quality of images as well as the ability to reproduce the desired attribute increases as the fraction of the labeled data grows. However, there is no significant improvement between $10\%$ and full supervision, which affirms the ability of our DMVAE to effectively use small amounts of paired (labeled) data.}

\label{fig:celeba_cross_prior}
\end{figure}

\begin{figure}[tbhp]
\centering
\begin{tabular}{>{\centering\arraybackslash}m{0.15\linewidth}>{\centering\arraybackslash}m{0.05\linewidth}>{\centering\arraybackslash}m{0.22\linewidth}>{\centering\arraybackslash}m{0.22\linewidth}>{\centering\arraybackslash}m{0.22\linewidth}>{\centering\arraybackslash}m{0.22\linewidth}}
Conditioning (style)      &
$x_2$ & 1\% Labels & 10\% Labels & 100\% Labels \\ 

&\rotatebox[origin=c]{90}{  Reconstruction}
&{\includegraphics[width=0.99\linewidth]{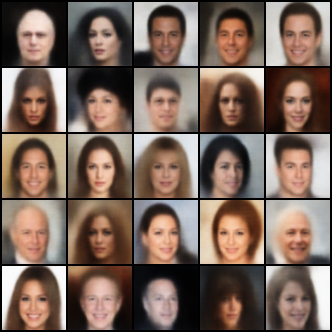}}
&{\includegraphics[width=0.99\linewidth]{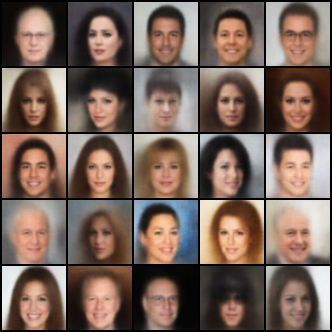}}
&{\includegraphics[width=0.99\linewidth]{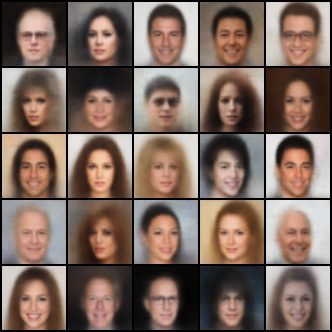}} \\ 

\multirow{4}{*}{
\begin{tabular}{c}
$x_1$ \\ 
\includegraphics[width=0.99\linewidth]{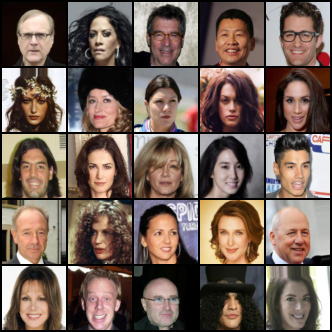}\\
\end{tabular}
}  &

\rotatebox[origin=c]{90}{Male} 
&{\includegraphics[width=0.99\linewidth]{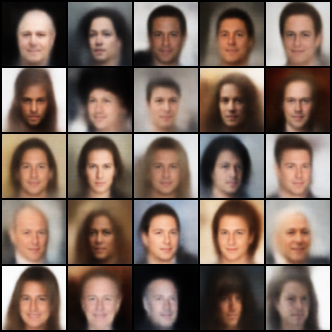}}
&{\includegraphics[width=0.99\linewidth]{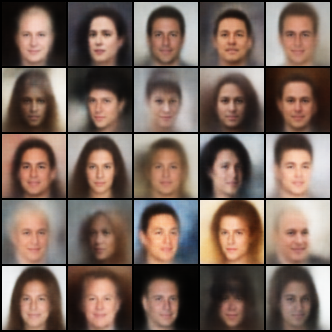}}
&{\includegraphics[width=0.99\linewidth]{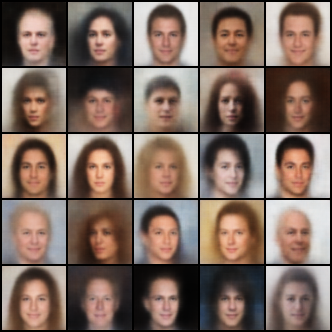}}\\

  & 
\rotatebox[origin=c]{90}{Smiling}  &
{\includegraphics[width=0.99\linewidth]{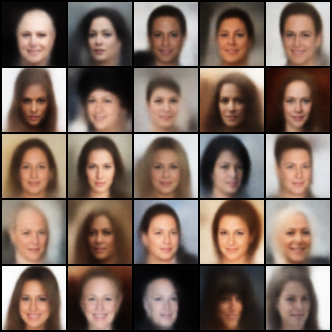}}&
{\includegraphics[width=0.99\linewidth]{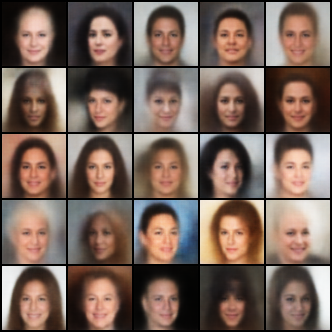}}&
{\includegraphics[width=0.99\linewidth]{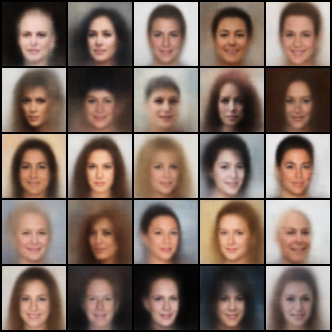}}\\ 

  & 
\rotatebox[origin=c]{90}{Eyeglasses}  &
{\includegraphics[width=0.99\linewidth]{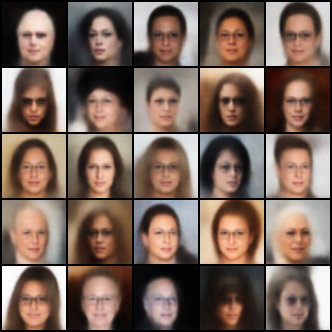}}&
{\includegraphics[width=0.99\linewidth]{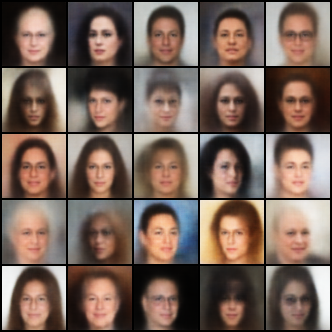}}&
{\includegraphics[width=0.99\linewidth]{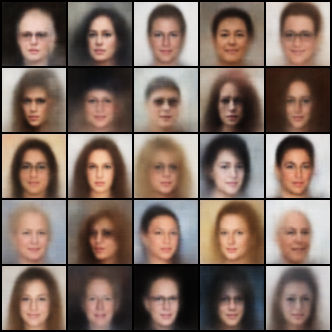}}\\ 
\end{tabular}
\caption{Cross-synthesis CelebA images from labels in full modality where $z_{p,Image}$ is from test image.  Rows of images correspond to reconstructed images, `Male', `Smiling', and `Eyeglasses', as depicted.  The columns correspond to DMVAE models trained on data with $1\%$, $10\%$, and $100\%$ labels.}
\label{fig:celeba_cross_test}
\end{figure}

\begin{figure}[tbhp]
\centering
\begin{tabular}{>{\centering\arraybackslash}m{0.15\linewidth}>{\centering\arraybackslash}m{0.05\linewidth}>{\centering\arraybackslash}m{0.72\linewidth}}

Conditioning (style)      &
& 
Cross-synthesis

\\ 
& & \includegraphics[width=0.99\linewidth]{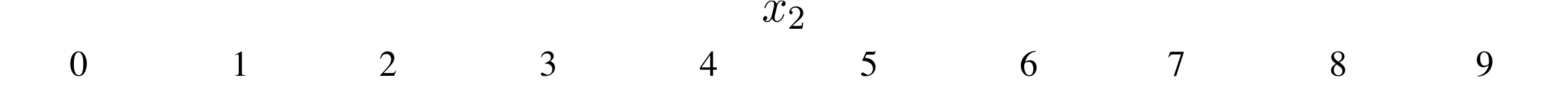}
\\

\multirow{2}{*}{
\begin{tabular}{c}
$x_1$\\
\includegraphics[width=0.7\linewidth]{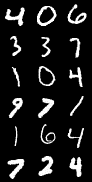}
\end{tabular}
}  

&\rotatebox[origin=l]{90}{0.2\%} 
&{\includegraphics[width=0.99\linewidth]{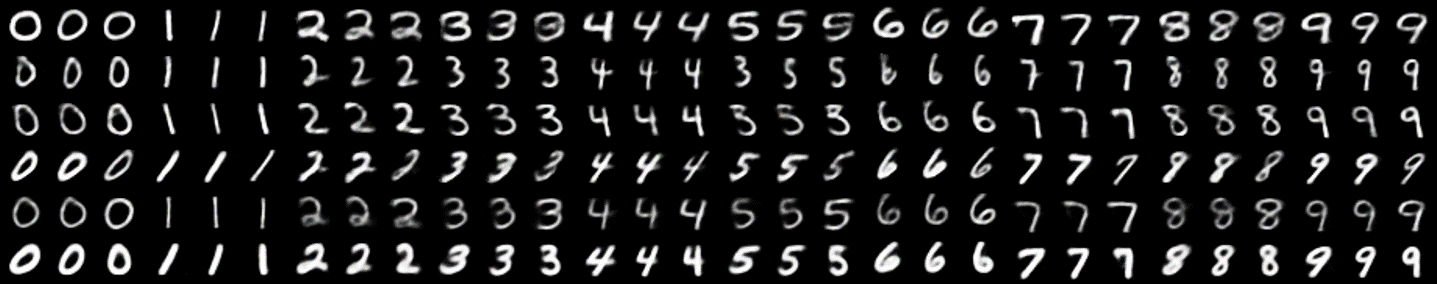}}\\

&\rotatebox[origin=l]{90}{100\%}  &
{\includegraphics[width=0.99\linewidth]{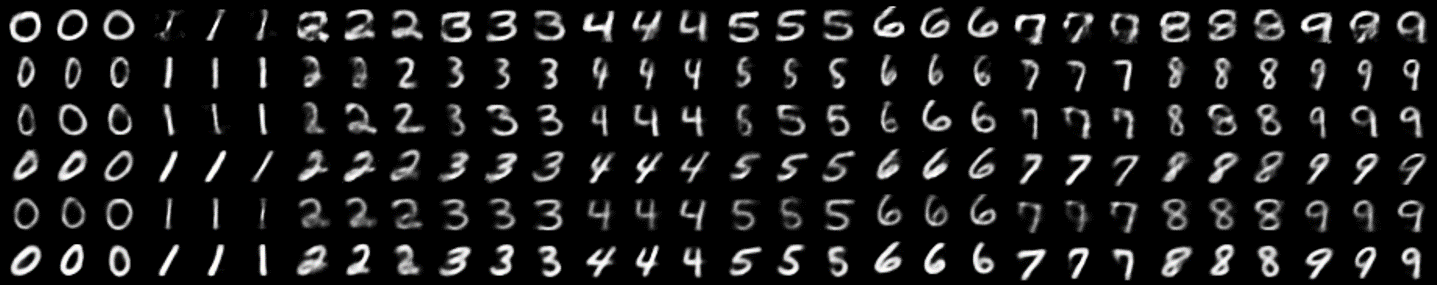}}\\ 

& & \includegraphics[width=0.99\linewidth]{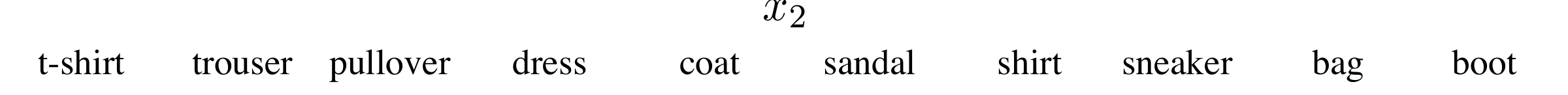}
\\

\multirow{2}{*}{
\begin{tabular}{c}
$x_1$\\
\includegraphics[width=0.7\linewidth]{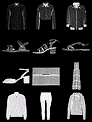}
\end{tabular}
}  

&\rotatebox[origin=l]{90}{0.2\%} 
&{\includegraphics[width=0.99\linewidth]{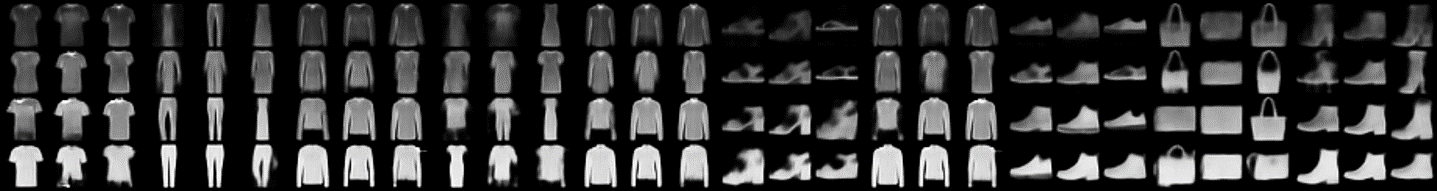}}\\

&\rotatebox[origin=l]{90}{100\%}  &
{\includegraphics[width=0.99\linewidth]{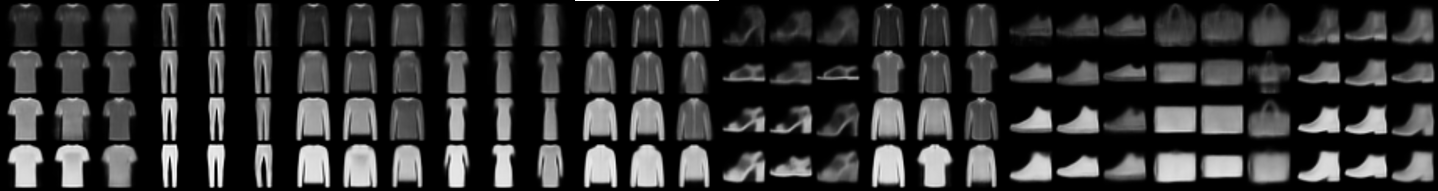}}\\ 

\end{tabular}
\caption{Cross-synthesis MNIST and FMNIST images from labels in full modality where $z_{p,Image}$ is from test image. Rows of images correspond to DMVAE models trained on data with $0.2\%$ and $100\%$ paired data.}
\label{fig:mnist_cross_test}
\end{figure}

\begin{figure}[tbhp]
\centering
\begin{tabular}{>{\centering\arraybackslash}m{0.15\linewidth}>{\centering\arraybackslash}m{0.05\linewidth}>{\centering\arraybackslash}m{0.72\linewidth}}

Conditioning (style)      &
& Cross-synthesis \\ 

& & \includegraphics[width=0.99\linewidth]{figures/x2zeronine.pdf}
\\

&\rotatebox[origin=l]{90}{0.2\%} 
&{\includegraphics[width=0.99\linewidth]{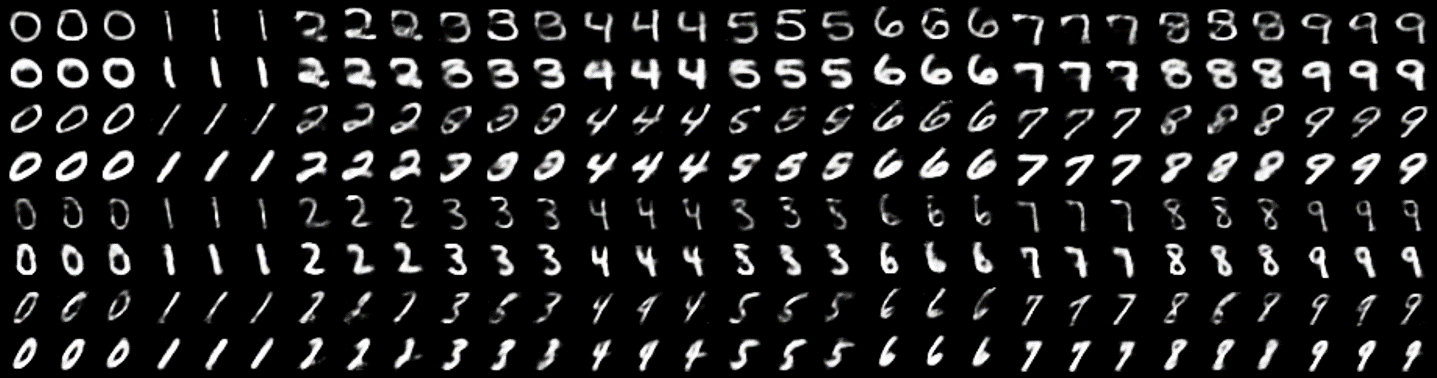}}\\

\pgfmathdeclarefunction{gauss}{2}{\pgfmathparse{1/(#2*sqrt(2*pi))*exp(-((x-#1)^2)/(2*#2^2))}}

\multirow{4}{*}{
\begin{tikzpicture}
\begin{axis}[width=3cm, tick style={draw=none}, xticklabels={},yticklabels={}]
\addplot [draw=black, domain=-3:3] {gauss(0,1)};
\end{axis}
\node[] at (.7,-.5) {$z_{p_1} \sim {\cal N}(0,1)$};
\end{tikzpicture}
}

&\rotatebox[origin=l]{90}{100\%}  &
{\includegraphics[width=0.99\linewidth]{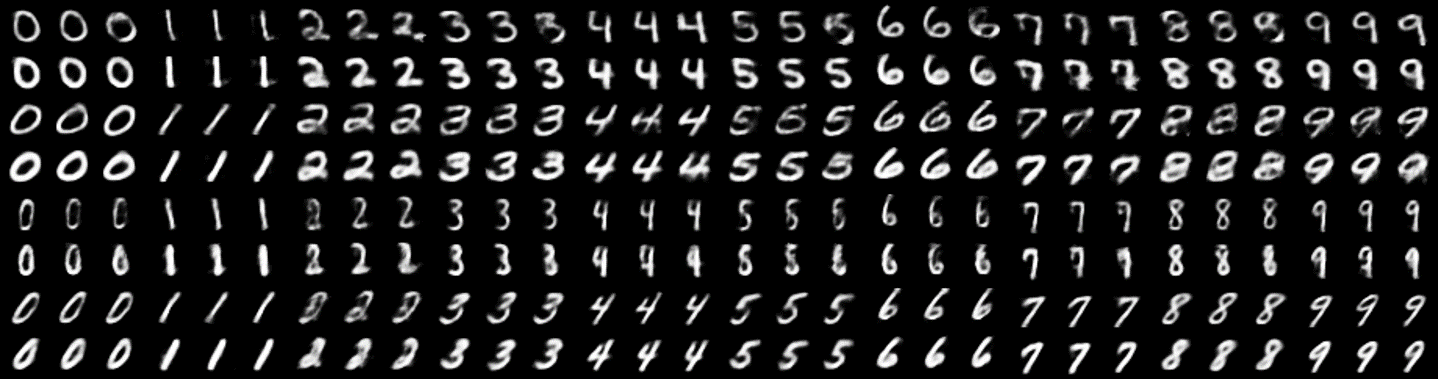}}\\ 

& & \includegraphics[width=0.99\linewidth]{figures/x2fashionlabels.pdf}
\\

&\rotatebox[origin=l]{90}{0.2\%} 
&{\includegraphics[width=0.99\linewidth]{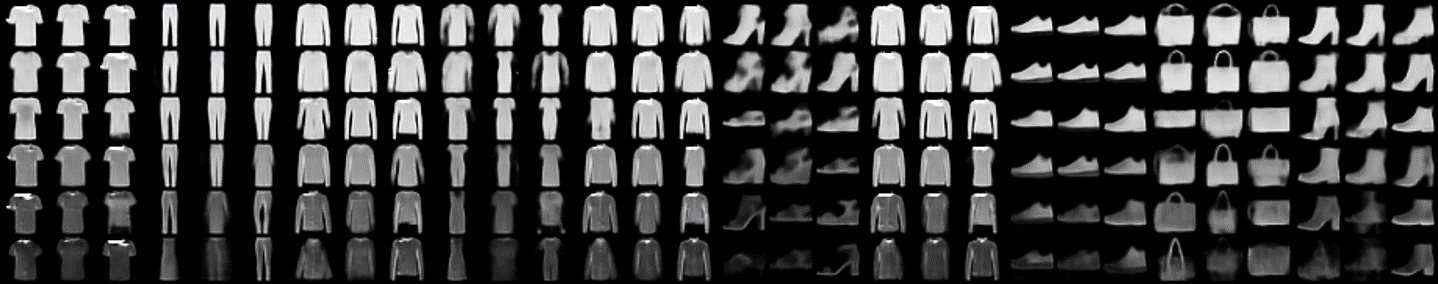}}\\

&\rotatebox[origin=l]{90}{100\%}  &
{\includegraphics[width=0.99\linewidth]{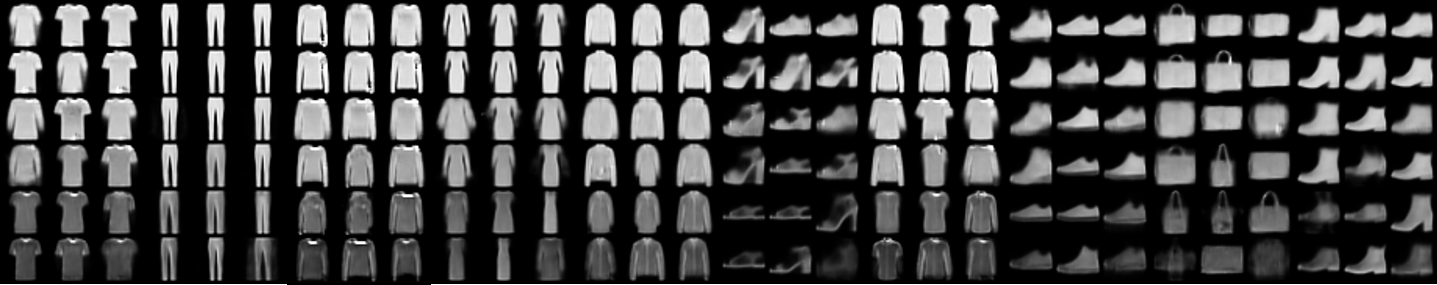}}\\ 

\end{tabular}
\caption{Cross-synthesis MNIST and FMNIST images from labels in full modality where $z_{p,Image}$ is $z_{p,Image} \sim {\cal N}(0,1)$. Rows of images correspond to DMVAE models trained on data with $0.2\%$ and $100\%$ paired data.}
\label{fig:mnist_cross_prior}
\end{figure}



\begin{figure}[tbhp]
\centering
\begin{tabular}{
>{\centering\arraybackslash}m{0.05\linewidth}
>{\centering\arraybackslash}m{0.37\linewidth}
>{\centering\arraybackslash}m{0.37\linewidth}
}
      & MNIST & FMNIST \\ 
\rotatebox[origin=c]{90}{0.2\%} & {\includegraphics[width=0.99\linewidth]{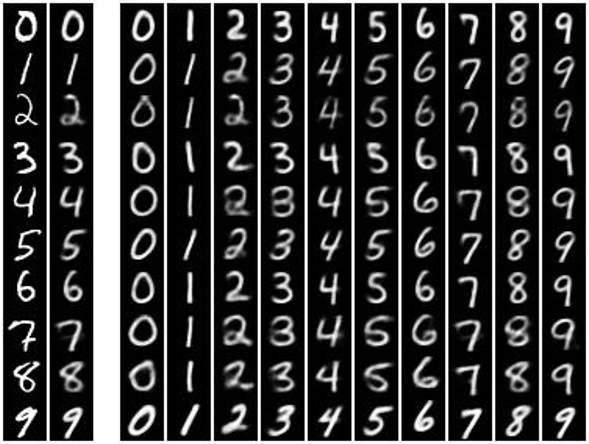}}
&{\includegraphics[width=0.99\linewidth]{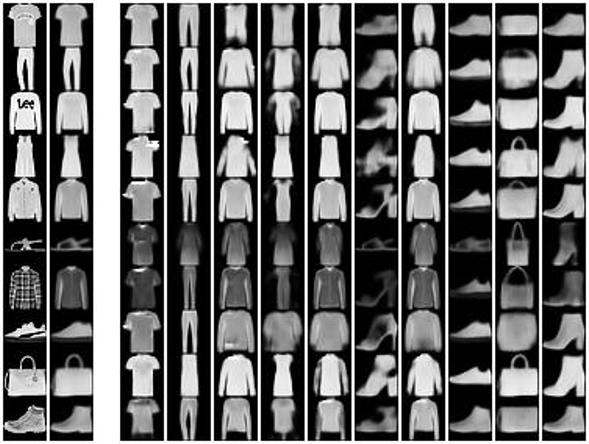}} \\ 

\rotatebox[origin=l]{90}{100\%}  &  {\includegraphics[width=0.99\linewidth]{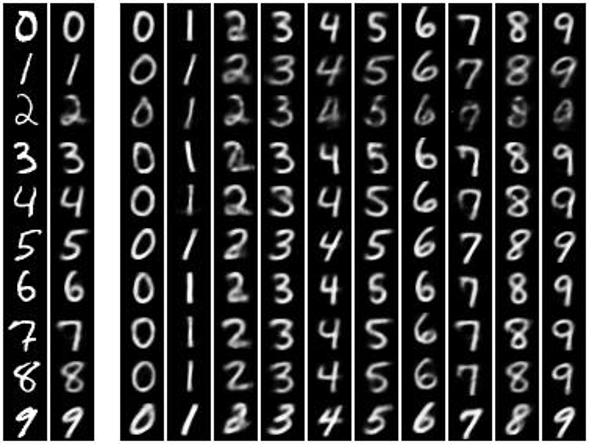}}
&{\includegraphics[width=0.99\linewidth]{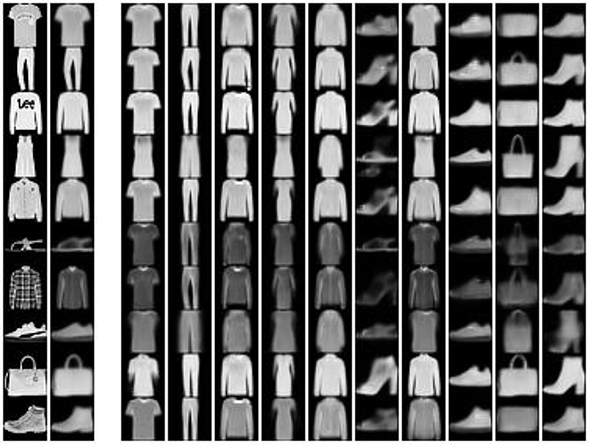}} \\ 

\end{tabular}
\caption{Shared space traversal experiment.  Style (private) latent variable is determined by the images in the first column (of the two-column set).  The second column in this set depicts the DMVAE resynthesis/reconstruction. The ten subsequent columns are the results of the traversal in the shared space. Because of the discrete nature of the shared space, the traversal can be seen as conditioning on all ten different digit, $\{0,\ldots,9\}$, or clothes types, $\{$T-shirt/top, $\dots$, Ankle boot$\}$ class labels.
 }
\label{tab:mnist_trv}
\end{figure}

\noindent\textbf{CelebA.} \autoref{fig:celeba_cross_prior} depicts cross-synthesis results with other attributes under missing modality, model \autoref{fig:cr_infer_missing}. Although the label is given as binary vector with possibly multiple activated attributes, we impose four instances of the label attribute: `Neutral', where all attributes are inactive; `Male', where the only active attribute is Male; `Smiling', where the only active attribute is smiling; and `Eyeglasses', where the active attribute is eyeglasses on. Cross-synthesized images conditioned on `Neutral' do not reveal attributes such as Smiling and Mouth Slightly Open, suggesting good separation of attributes in the shared latent space. However, inactive `Male' attribute depicts `Female' faces, implying dependency between it and the 'Male' attribute. In contrast, the cross-synthesis from `Male' or `Smiling' consistently changes the attribute of all faces to male or smiling respectively.  Finally, in the case of `Eyeglasses', generating images with this attribute is more challenging. This is due to severe imbalance of training examples with this attribute, as shown in the \autoref{fig:celeba_stat}.
While the quality of synthesized results deteriorates as the label fraction gets lower, as expected, our model is still able to synthesize reasonable results with small amounts of labeled data.

\autoref{fig:celeba_cross_test} shows the cross-synthesis results where $z_{p,image}$ is obtained from the conditioning images, similar to Fig.8 in the main paper. The Reconstruction case corresponds to the instance where the imposed attributes correspond to the actual attributes in the conditioning images, i.e., the model depicted in \autoref{fig:cr_infer_complete}. The last three rows are the instances where the attributes of the conditioning images are changed into the attributes `Male', `Smile', and `Eyeglasses', the model in \autoref{fig:cr_infer_cond}.
We note that the use of specific style from the conditioning image enables synthesis of realistic reconstructions even using the model trained with only 1\% labeled data.

\noindent\textbf{Digits.}  In  \autoref{fig:mnist_cross_test} and \autoref{fig:mnist_cross_prior}, we depict the cross-synthesis results of MNIST and FMNIST using the conditioning reconstruction defined in  \autoref{fig:cr_infer_cond} and the missing modality in reconstruction defined in \autoref{fig:cr_infer_missing} respectively, in addition to the results in Fig. 2 in the main paper.

In  \autoref{fig:mnist_cross_test}, we pick the test samples that can represent visually distinct styles as displayed on the left hand side. In 6 $\times$ 3 block of MNIST test images, each row shows (wide, narrow, not-slanted, slanted, thin, thick) style. In 4 $\times$ 3 block of FMNIST test images, the brightness is varied. On the right hand side, the conditioning reconstruction of each class is well-generated representing the target class. Moreover, the styles of conditioning test images are kept at the same time.

In \autoref{fig:mnist_cross_prior}, for MNIST, we give extreme values to the private factors obtained for (width, slant, thickness) after sampling $z_{p_1}$ from the prior distribution as described in the main paper. For FMNIST, we vary the private factors obtained for brightness in six-range, [-3,-2,1,0,1,2].
8 $\times$ 3 block of MNIST images and 6 $\times$ 3 block of FMNIST images are generated from the same conditioning label. We note that each of classes are realistically synthesized, while preserving the intended style (private) factors, rather it is sampled randomly from the prior distribution of $z_{p_1}$.

Next, we analyze the impact of each latent factor on the generation of an image. \autoref{tab:mnist_trv} depicts the results of varying (traversing) the discrete shared latent factor for MNIST and FMNIST. The first two columns are the ground-truth image and the reconstructed image, in order. In the subsequent ten columns, each column shows the generated images when a different one-hot vector of 10-dimensions is fed as the shared latent feature,  while the private latent feature (style) of the original image is fixed. In MNIST, we see that the style of the digits remains preserved, while the content varies independently of that style.  This suggests the information about digit identity exists only in the shared space, separated from the private style space. 
However, in FMNIST, when 0.2\% paired data is used, the tops are not separated clearly in the shared space, which results in only 0.749 accuracy as reported in the main paper.

\begin{figure}[tbhp]
\centering
\begin{tabular}{
>{\centering\arraybackslash}m{0.05\linewidth}
>{\centering\arraybackslash}m{0.3\linewidth}
>{\centering\arraybackslash}m{0.3\linewidth}
>{\centering\arraybackslash}m{0.3\linewidth}
}
      & DMVAE-Private & DMVAE-Shared & 
      MVAE \\ 
\rotatebox[origin=c]{90}{Male} & {\includegraphics[width=0.99\linewidth]{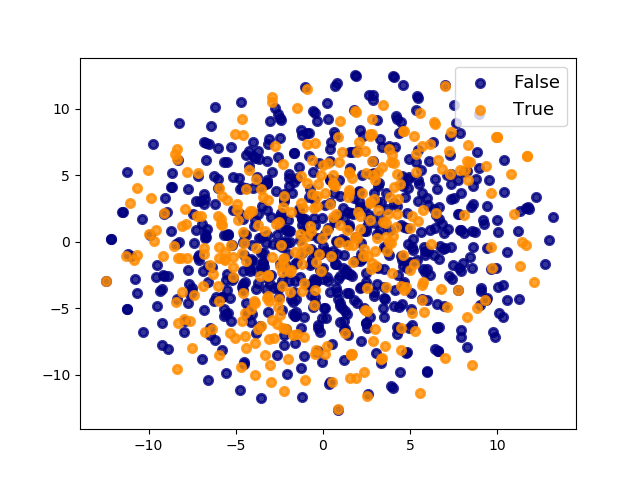}}
&{\includegraphics[width=0.99\linewidth]{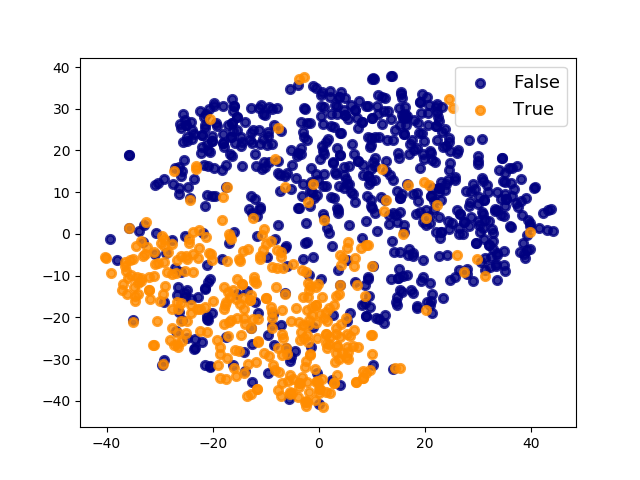}}
&{\includegraphics[width=0.99\linewidth]{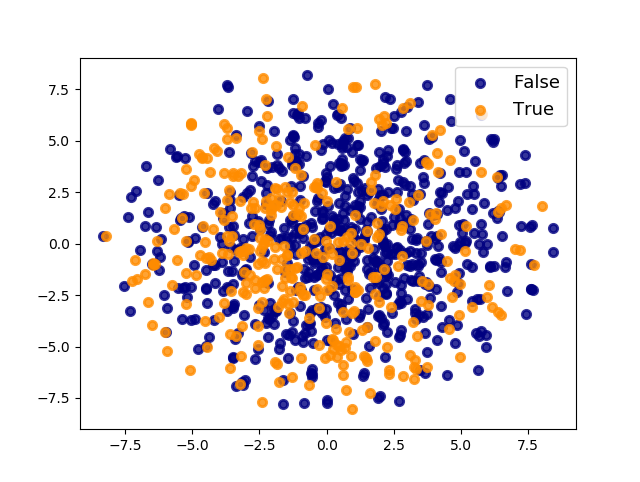}} \\ 

\rotatebox[origin=l]{90}{Heavy Makeup}  &  {\includegraphics[width=0.99\linewidth]{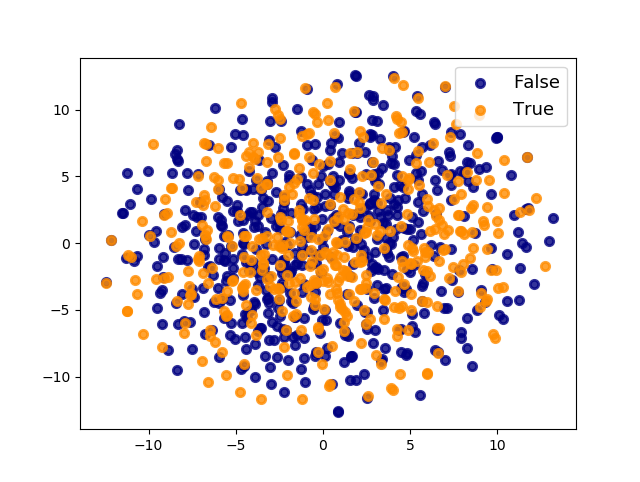}}& 
{\includegraphics[width=0.99\linewidth]{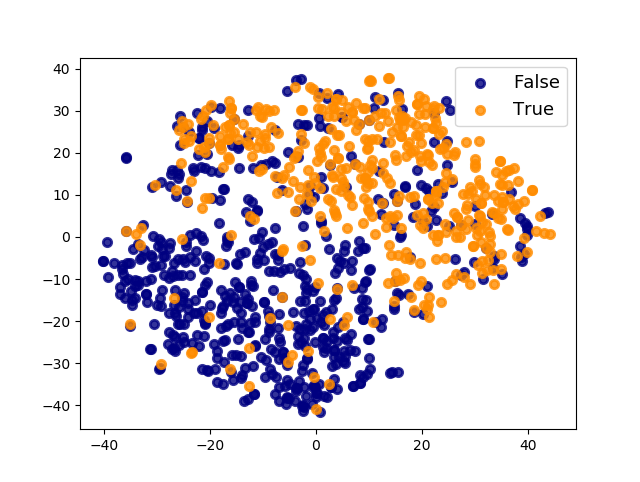}}&
{\includegraphics[width=0.99\linewidth]{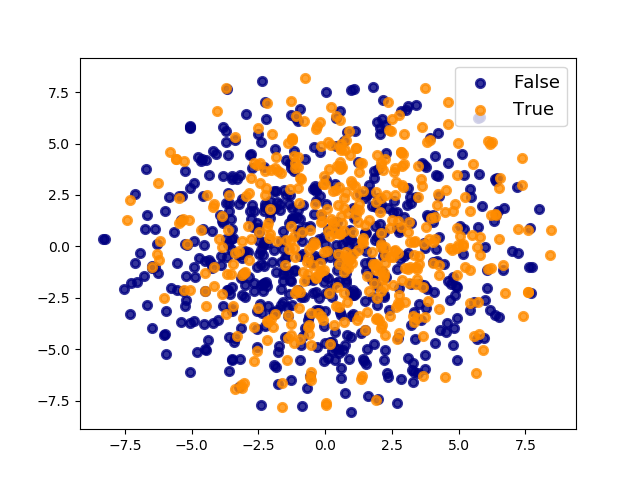}}\\ 
\end{tabular}
\caption{ 
  Visualization of 2-D embeddings of shared $z_s$ and private $z_{p,Image}$ features for DMVAE and latent features $z$ for MVAE.  Each row corresponds to a single attribute (`Male' and `Heavy Makeup') of the data set to a fixed value (points with attribute = False are colored orange, while those with attribute = True are colored blue).  Note that DMVAE-Shared features are, as expected, largely correlated with the attribute value, more strongly for `Male' and less so for `Heavy Makeup', indicating the ability of our DMVAE to learn and represent that value in the shared space.  The features in the DMVAE-Private space are, also as expected, not correlated with the attribute values, suggesting that DMVAE-Private contains information independent of the attribute in question and private to the image space (style).  However, MVAE is unable to learn the representation of the fixed attribute in the latent space, as indicated by the absence of correlation of the embeddings with the attribute False/True values.}
\label{tab:dmvae_celeba_embed}
\end{figure}

\subsection{Separation of Private and Shared Latent Spaces} \label{sec3.sepa}

We investigate the separation of information across private and shared latent spaces by projecting the latent features inferred by the encoders, on the test set, into a 2-D space with tSNE. We use 1,000 randomly selected samples to plot the embedded features.

\noindent\textbf{CelebA.}. We extract 1,000 features under our DMVAE model trained with 1\% of the total attribute labels. The embedding is illustrated in \autoref{tab:dmvae_celeba_embed}. Since CelebA data has multiple attributes, we focus on the embedded features restricted to a single attribute, either `Male' or `Heavy Makeup'. (`Male', DMVAE-Shared) shows how the shared features $z_s$ embedded into a 2-D space, where the data points are colored according to the presence or absence of the `Male' attribute. The separation of the two groups is evident even in this low-dimensional embedding.  However, (`Male', DMVAE-Private), the embedding of $z_{p,Image}$, does not reflect any dependency of the attribute value with the features.  This suggests that our DMVAE model learns to place the `Male' attribute into the shared space, whose dimensions (horizontal in the embedded space) largely reflects the strength of that attribute. On the other hand, the attribute-irrelevant image style information remains in the private space.  In contrast, in (`Male', MVAE) there is no clear disentanglement of the factor `Male' nor its association with a feature of that space.  Similar conclusions can be drawn for the with respect to `Heavy Makeup' attribute.

\begin{figure}[tbhp]
\centering
\begin{tabular}{
>{\centering\arraybackslash}m{0.05\linewidth}
>{\centering\arraybackslash}m{0.3\linewidth}
>{\centering\arraybackslash}m{0.3\linewidth}
>{\centering\arraybackslash}m{0.3\linewidth}
}
      & DMVAE-Private & DMVAE-Shared & 
      MVAE \\ 
\rotatebox[origin=l]{90}{MNIST} & {\includegraphics[width=0.99\linewidth]{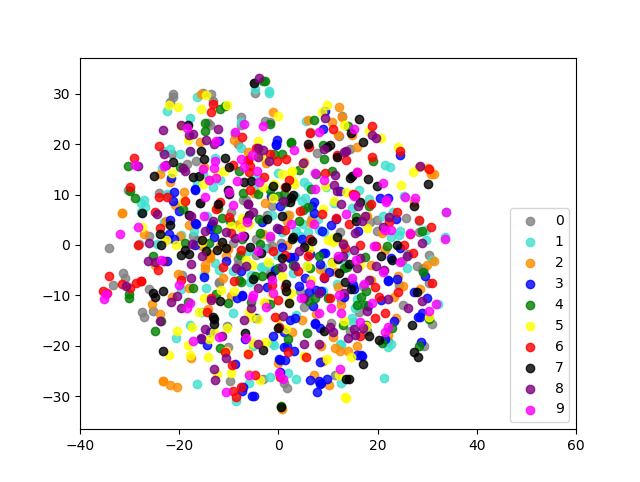}}
&{\includegraphics[width=0.99\linewidth]{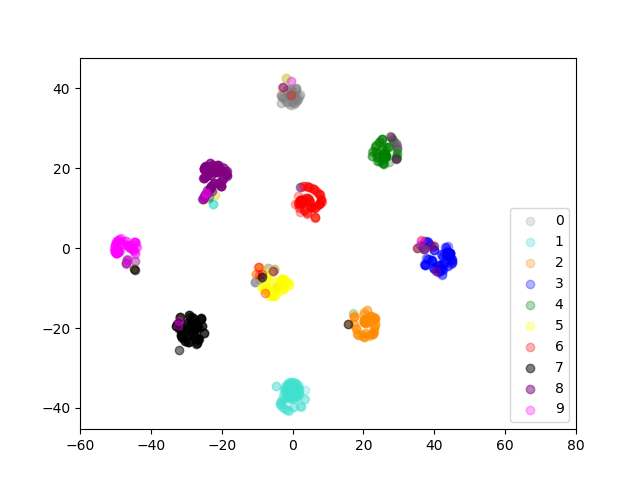}}
&{\includegraphics[width=0.99\linewidth]{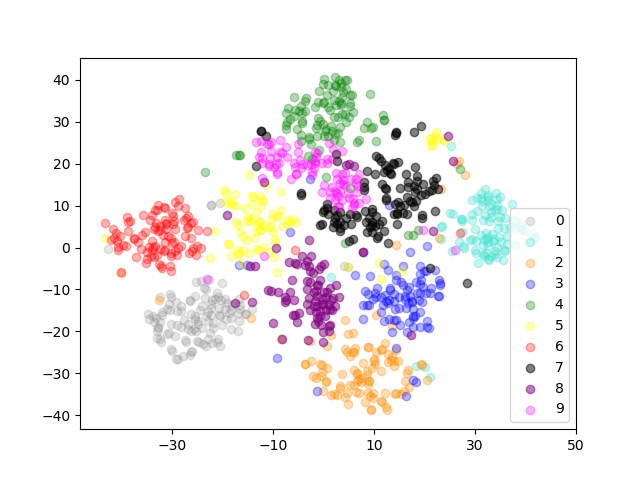}} \\ 

\rotatebox[origin=l]{90}{FMNIST}  &  {\includegraphics[width=0.99\linewidth]{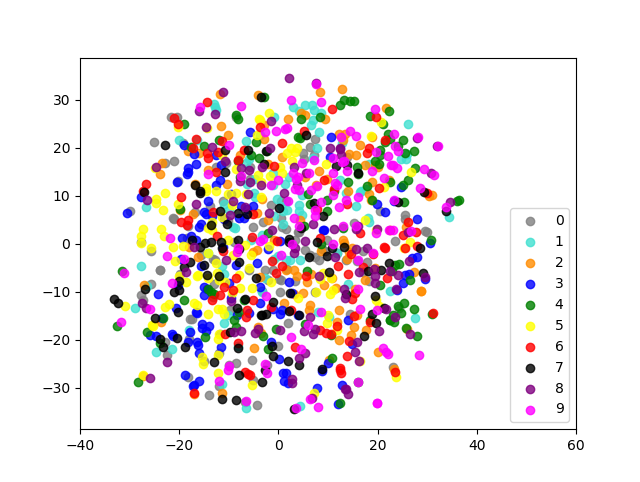}}
&{\includegraphics[width=0.99\linewidth]{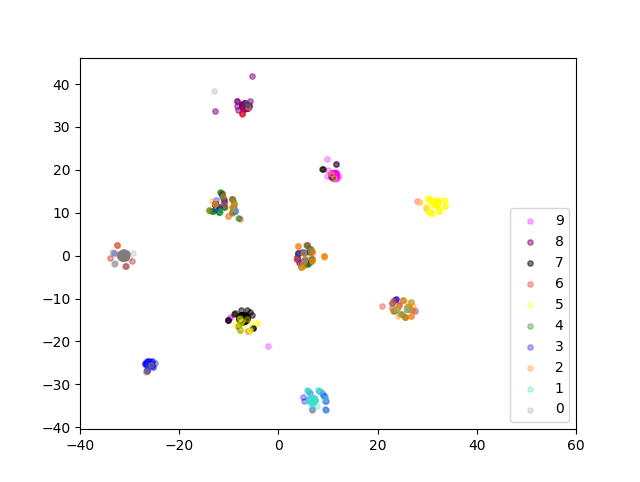}}
&{\includegraphics[width=0.99\linewidth]{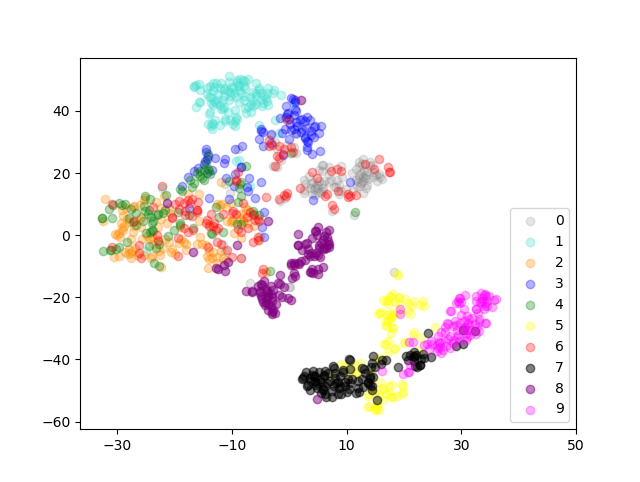}}\\ 
\end{tabular}
\caption{
  Visualization of 2-D embeddings of shared $z_s$ and private $z_{p,Image}$ features for DMVAE and latent features $z$ for MVAE using tSNE.
Each color associates with one of ten digit classes, $\{0,1,...,9\}$.}
\label{fig:dmvae_mnist_embed}
\end{figure}

\noindent\textbf{Digits.} For MNIST and FMNIST, 1,000 features are extracted with the model trained with 0.2\% paired data. 
\autoref{fig:dmvae_mnist_embed} shows that within private factors, the data points are scattered randomly in the embedded 2-D space, suggesting the lack of dependency (as desired) between the Private space features (image style) and the class labels (image content).  On the other hand, the Shared space features associate well with the classes $\{0,1,...,9\}$. 
Though MVAE shows reasonable separation of each class, the embedding of each class exist nearby each other and overlapped, which causes the confusion of the model to predict the class label.

\subsection{Published and Reproduced Results}\label{sec:discrepancy}

In this section we highlight the discrepancy between the results reported in \cite{NIPS2018_7801} and those that can be reproduced using the authors' publicly available code at https://github.com/mhw32/multimodal-vae-public.  Both sets of results are reported in \autoref{tab:mnist}.  Every effort was made to replicate the published results, including contacting the authors and the extensive hyperparameter cross-validation beyond those reported by the authors.  However, the discrepancy in performance remained.  

\begin{table}[tbhp]
\caption{Classification accuracy for MNIST and FMNIST as a function of the paired data fraction (in \% points). MVAE\cite{NIPS2018_7801}  represents the results published in~\cite{NIPS2018_7801}, while MVAE stands for the reproduced results based on the authors' code, which we show in Tab. 1 in the main paper. Bold denotes the highest accuracy among MVAE and our DMVAE. Red denotes the best score among MVAE, MVAE\cite{NIPS2018_7801}, and our DMVAE.  Finally, bold red indicates that the result is best according to both ``bold'' and red ``criteria''.  Note that the published results in~\cite{NIPS2018_7801} deviate from those reproduced in the first row of this table.}
\label{tab:mnist}
\begin{center}
\scriptsize

\setlength\tabcolsep{2.5pt}
\begin{tabular}
{lN{1}{3}N{1}{3}N{1}{3}N{1}{3}N{1}{3}N{1}{3}N{1}{3}N{1}{3}
N{2}{1}
N{1}{3}N{1}{3}N{1}{3}N{1}{3}N{1}{3}N{1}{3}N{1}{3}N{1}{3}
}

\toprule
 Model & \multicolumn{8}{c}{MNIST} & \multicolumn{9}{c}{FMNIST} \\
\midrule
  & {0.1}  & {0.2}  & {0.5}  & {{1}}   & {{5}}    & {10}   & {50}  & {100}
&& {0.1}  & {0.2}  & {0.5}  & {{1}}   & {{5}}    & {10}   & {50}  & {100}
  \\ 
\midrule

MVAE  
& 0.4409 & 0.5793 & 0.829 & 0.8967 & 0.935 & 0.9481 & 0.9613 & 0.9711
&& 0.4865 & 0.5591 & 0.66 & 0.7725 & 0.8265 &	0.8423 & 0.8692 & 0.882
\\
MVAE\cite{NIPS2018_7801} 
& 0.2842 & 0.6254 & 0.8593 & 0.8838 & \topRed 0.9584 & \topRed 0.9711 & 0.9678 & 0.9681 
&& 0.4548 & 0.5189 & 0.7619 & \topRed 0.8619 & \topRed 0.9243 & \topRed 0.9239 & \topRed 0.9478 & \topRed 0.947  \\

\midrule
DMVAE  & \topBRed 0.7735 & \topBRed 0.9166 & \topBRed 0.9443 & \topBRed 0.9455 & \topscore 0.9573 & \topscore 0.9605 & \topBRed 0.9740 & \topBRed 0.9810 
&& \topBRed 0.5406 & \topBRed 0.7490 & \topBRed 0.7774 & \topscore 0.8335 &  \topscore 0.8709 & \topscore 0.8937 & \topscore 0.9280 & \topscore 0.941  \\

\bottomrule                    
\end{tabular}
\end{center}
\end{table}

\section{Ablation study}
We investigated the effectiveness of each component of DMVAE on MNIST in Sec. 5.3 of the main paper. We further check the importance of separated private and shared space as well as the discrepancy coming from hybrid model on FMNIST and CelebA. \autoref{tab:ablation} shows all DMVAE settings provide better performance than MVAE except for the CelebA accuracy. Within DMVAE, discrete factors improve prediction as they are better aligned with the categorical nature of attributes.

\begin{table}[tbhp]
\caption{Ablation study to compare MVAE, DMVAE continuous shared space, and DMVAE discrete shared space.}
\label{tab:ablation}
\begin{center}

\setlength\tabcolsep{2.5pt}
\begin{tabular}
{llN{1}{3}N{1}{3}N{1}{3}N{1}{3}}
\toprule
  & Metric   & {MVAE}  & {DMVAE\_cont}  & {DMVAE\_disc} \\ \midrule
MNIST & Accuracy & 0.625 & 0.695  &  \topscore 0.917  \\
FMNIST  & Accuracy & 0.559  & 0.631 & \topscore 0.749 \\
 & F1-score &0.444 & 0.502 & \topscore 0.522 \\
\multirow{-2}{*}{CelebA} & Accuracy & \topscore 0.863 & 0.858 &  0.862	      \\
\bottomrule                    
\end{tabular}
\end{center}
\end{table}

\end{document}